\documentclass{article}
     \PassOptionsToPackage{numbers, sort&compress}{natbib}
\usepackage[final]{neurips_2019}
\usepackage[utf8]{inputenc} \usepackage[T1]{fontenc}    \usepackage{hyperref}       \usepackage{url}            \usepackage{booktabs}       \usepackage{amsfonts}       \usepackage{nicefrac}       \usepackage{microtype}      \usepackage{floatrow}
\usepackage[para]{footmisc}
\newfloatcommand{capbtabbox}{table}[][\FBwidth]
\usepackage{caption}
\usepackage{bm}
\usepackage{nicefrac}
\usepackage{graphicx}
\usepackage{booktabs} \usepackage{array}
\usepackage{comment}
\usepackage{tikz}
\usepackage{subfig}
\usepackage{amsmath, amssymb, amsthm}
\usepackage{mathtools}
\usepackage{xspace}
\usepackage{svg}
\usepackage{pifont}
\usepackage{algorithm}
\usepackage{algorithmic}
\usepackage{float}

\def\bea{\begin{eqnarray}}
\def\eea{\end{eqnarray}}
\def\fig#1{Figure~\ref{fig:#1}}
\def\tab#1{Table~\ref{tab:#1}}
\def\sect#1{Section~\ref{sec:#1}}

\def\app#1{Appendix~\ref{app:#1}}

\makeatletter
\DeclareRobustCommand\onedot{\futurelet\@let@token\@onedot}
\def\@onedot{\ifx\@let@token.\else.\null\fi\xspace}
\def\eg{\emph{e.g}\onedot} 
\def\ie{\emph{i.e}\onedot} 
 
 \def\vs{\emph{vs}\onedot} 
\def\wrt{w.r.t\onedot}

\makeatother

\def\figvspaceOne{\vspace{-3mm}}
\def\figvspaceTwo{\vspace{-5mm}}
\def\mypar#1{{\noindent\bf #1.}\hspace{1mm}}

\newcommand{\kld}[2]{\ensuremath{\mathcal{D}_{\mathrm{KL}}(#1||#2)}}

\newcommand{\norm}[1]{\left\lVert#1\right\rVert}
\DeclareMathOperator{\tr}{Tr}

\DeclareMathOperator{\JSD}{\mathcal{D}_{\textrm{JS}}}
\DeclareMathOperator*{\mean}{\mathbb{E}}

\title{Adaptive Density Estimation for Generative Models}

\author{	Thomas Lucas$^{*}$ \\   Inria$^{\dagger}$\\
  \And
  Konstantin Shmelkov$^{*,\ddagger}$  \\
  Noah's Ark Lab, Huawei \\
     \AND
  Cordelia Schmid\\
  Inria$^{\dagger}$\\
  \And
  Karteek Alahari\\
  Inria$^{\dagger}$\\
  \And
  Jakob Verbeek\\
  Inria$^{\dagger}$\\
}
      
\begin{document}
\maketitle

\newcommand{\Z}{\mathcal{Z}}
\newcommand{\red}[1]{\textcolor{red}{#1}}
\definecolor{burntorange}{rgb}{0.8, 0.33, 0.0}
\newcommand{\ora}[1]{\textcolor{burntorange}{#1}}
\newcommand{\gre}[1]{\textcolor{green}{#1}}
\newcommand{\ble}[1]{\textcolor{blue}{#1}}
\newcommand{\mcheck}{\checkmark}
\newcommand{\nocheck}{$ $}

\newcommand{\q}{p_{\theta}}
\newcommand{\p}{p^{*}}

\newcommand{\mQ}{\mathcal{Q}}
\newcommand{\mP}{\mathcal{P}}
\newcommand{\sqe}{R_{Q}^{\epsilon}}

\newcommand{\PEX}{\mathcal{P}^{\epsilon}(\mathcal{X})}
\newcommand{\PX}{\mathcal{P}(\mathcal{X})}

\newcommand{\RP}{R_{p}}
\newcommand{\REP}{R_{\p}}
\newcommand{\REQ}{R_{\q}}
\newcommand{\DFDX}{\frac{\partial f_{\psi}(x)}{\partial x^\top}}

\newcommand{\SC}{s_c}
\newcommand{\enc}{q_\phi}
\newcommand{\CIFAR}{CIFAR-10}

\newcommand{\Y}{\checkmark}
\newcommand{\victory}{\checkmark}
\newcommand{\N}{$\times$}
\newcommand{\CQFG}{AV-ADE\xspace}
\newcommand{\CQG}{AV-GDE\xspace}
\newcommand{\piVAE}{V-ADE\xspace}

\begin{abstract}
Unsupervised learning of generative models has seen tremendous progress over recent years,
in particular due to generative adversarial networks (GANs), variational autoencoders, and flow-based models.
GANs have dramatically improved sample quality, but suffer from two drawbacks:
(i) they mode-drop, \ie,  do not cover the full support of the train data, 
and (ii) they do not allow for likelihood evaluations on held-out data.
In contrast, likelihood-based training encourages models to cover the full support of the train data, but yields poorer samples.
These mutual shortcomings can in principle be addressed by training generative latent variable models in a hybrid adversarial-likelihood manner.
However, we show that commonly made parametric assumptions create a conflict between them, making successful hybrid models non trivial.
As a solution, we propose to use deep invertible transformations in the latent variable decoder. 
This approach allows for likelihood computations in image space, is more efficient than fully invertible models, and can take full advantage of adversarial training.
We show that our model significantly improves over existing hybrid models: offering GAN-like samples, IS and FID scores that are competitive with fully adversarial models, and improved likelihood scores. 
\end{abstract}

\newenvironment{splitfigure}[3]
	{\begin{figure}
		\begin{minipage}[c]{0.59\textwidth}
		#1
		\end{minipage}\hfill
		\begin{minipage}[c]{0.39\textwidth}
			\addtocounter{figure}{1}
					Figure \arabic{figure}: #2	
		\end{minipage}
				\label{fig:cdt_qdt}
         \end{figure}}

\newcommand{\baselinesFig}{
	\begin{figure}
		\begin{minipage}[c]{0.59\textwidth}
				\begin{tabular}{ccc}
					\hspace*{-.5cm} \setlength{\tabcolsep}{0pt}
\begin{tabular}{ccc} 
\includegraphics{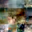} &
\includegraphics{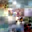} &
\includegraphics{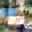}\\[-0.85ex]
\includegraphics{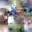} &
\includegraphics{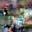} &
\includegraphics{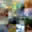}\\[-0.85ex]
\includegraphics{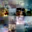} &
\includegraphics{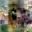} &
\includegraphics{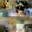}\\

\end{tabular} 
 & \hspace*{-0.20cm} \setlength{\tabcolsep}{0pt}
\begin{tabular}{ccc} 
\includegraphics{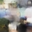} &
\includegraphics{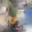} &
\includegraphics{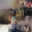}\\[-0.85ex]
\includegraphics{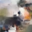} &
\includegraphics{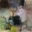} &
\includegraphics{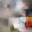}\\[-0.85ex]
\includegraphics{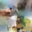} &
\includegraphics{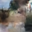} &
\includegraphics{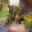}\\
\end{tabular} 
 & \hspace*{-0.20cm} \setlength{\tabcolsep}{0pt}
\begin{tabular}{ccc} 
\includegraphics{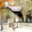} &
\includegraphics{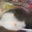} &
\includegraphics{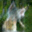}\\[-0.85ex]
\includegraphics{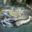} &
\includegraphics{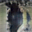} &
\includegraphics{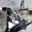}\\[-0.85ex]
\includegraphics{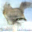} &
\includegraphics{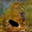} &
\includegraphics{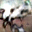}\\
\end{tabular} 
 \\
					VAE & \piVAE & \CQG \\
				\end{tabular} \\
				\begin{tabular}{cc}
					\hspace*{-0.5cm} \setlength{\tabcolsep}{0pt}
\begin{tabular}{ccccc} 
\includegraphics{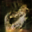} &
\includegraphics{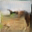} &
\includegraphics{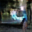} &
\includegraphics{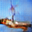} &
\includegraphics{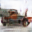}\\[-0.85ex]
\includegraphics{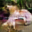} &
\includegraphics{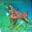} &
\includegraphics{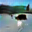} &
\includegraphics{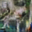} &
\includegraphics{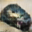}\\[-0.85ex]
\includegraphics{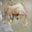} &
\includegraphics{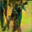} &
\includegraphics{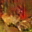} &
\includegraphics{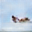} &
\includegraphics{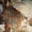}\\
\end{tabular} 
  &
					\hspace*{-0.5cm} \setlength{\tabcolsep}{0pt}
\begin{tabular}{ccccc} 
\includegraphics{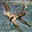} &
\includegraphics{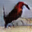} &
\includegraphics{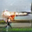} &
\includegraphics{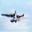} &
\includegraphics{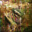}\\[-0.85ex]
\includegraphics{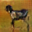} &
\includegraphics{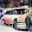} &
\includegraphics{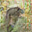} &
\includegraphics{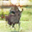} &
\includegraphics{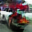}\\[-0.85ex]
\includegraphics{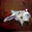} &
\includegraphics{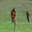} &
\includegraphics{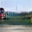} &
\includegraphics{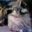} &
\includegraphics{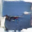}\\
\end{tabular} 
 \\
					 GAN & \CQFG (Ours)\\
				\end{tabular}
				\figvspaceOne 
		\end{minipage} \hfill
		\begin{minipage}[c]{0.39\textwidth}
			\setlength{\tabcolsep}{2pt}
			\footnotesize
			\begin{tabular}{lccccccr}
				\toprule
				& $f_{\psi}$ & Adv. & {\scriptsize MLE} & {\scriptsize BPD} $\downarrow$ & {\scriptsize IS} $\uparrow$ & {\scriptsize FID} $\downarrow$\\ 				\midrule
								{\scriptsize GAN}                                                                                        & \N & \Y           & \N            & [7.0]                & 6.8           & 31.4  \\
								{\scriptsize VAE}                                                                                        & \N & \N           & \Y            & 4.4              & 2.0           & 171.0 \\

				{\scriptsize \piVAE}$^{\dagger}$                                                                                       &\Y & \N           & \Y           & 3.5       & 3.0          & 112.0   \\
				\midrule
								{\scriptsize \CQG}  & \N & \victory           & \victory                   & 4.4    & 5.1  & 58.6 \\
								{\scriptsize \CQFG}$^{\dagger}$                                                                                        & \Y & \victory           & \victory           & 3.9              & 7.1           & 28.0   \\
				\bottomrule
			\end{tabular}
					
					\footnotesize
					\addtocounter{figure}{1}
					\addtocounter{table}{1}
					\vspace*{0.2cm}
					Table \arabic{table}: Quantitative results. $^{\dagger}$ : Parameter count decreased by $1.4\%$ to compensate for $f_{\psi}$.
					[Square brackets] denote that the value is approximated, see \sect{evaluation}.\\

					Figure \arabic{figure}: Samples from GAN and VAE baselines, our \piVAE,  \CQG and \CQFG models, all trained on CIFAR-10.
												\end{minipage}
	\end{figure}
	\def \tabBaselineCount {\arabic{table}}
	\def \figBaselineCount {\arabic{figure}}
		\label{fig:BaselineSamples}
	\label{tab:baselines} 
}

\newcommand{\hybridTable}{
	{
		\setlength{\tabcolsep}{3pt}
				{\renewcommand{\arraystretch}{1.0}
		\begin{tabular}[t]{lccc}
		 \toprule
		  Hybrid (L) & BPD $\downarrow$     & IS $\uparrow$ & FID $\downarrow$\\
									\midrule
		  \CQFG (wg, rd)          & $3.8$   &  {\bf 8.2}     &  {\bf 17.2} \\
		  \CQFG (iaf, rd)           & $3.7$   &  $8.1$     &  $18.6$ \\
		  \CQFG (S2)          & {\bf 3.5} &  $6.9$ &  $28.9$ \\
		  FlowGan(A)~\cite{flowgan}  & $8.5$ & $5.8$ &  \\
		  FlowGan(H)~\cite{flowgan}  & $4.2$ & $3.9$ & \\
		  		  			\bottomrule 
		\end{tabular}
		}
		\label{tab:sota_hybrid}
	}
}

\newcommand{\advHybridTable}{
	{
	\setlength{\tabcolsep}{3.5pt}
	\begin{tabular}[t]{lccc}
	  \toprule
	  Hybrid (A) \hspace{0.48cm}  & BPD $\downarrow$     & IS $\uparrow$ & FID $\downarrow$\\
	   \midrule
	  AGE~\cite{ulyanov18aaai} &        & $5.9$    &  \\
	  ALI~\cite{dumoulin17iclr} &        & $5.3$   &   \\
	  SVAE~\cite{symVAE} &        & $6.8$    &  \\
	  $\alpha\text{-GAN}$~\cite{rosca2017variational} &        & $6.8$    & \\
	   SVAE-r~\cite{symVAE} &        & {\bf 7.0}  &  \\
	   \bottomrule 
	\end{tabular}
	\label{tab:sota_hybrid_adv}
	}
}

\newcommand{\AdvTable}{
	{
		\setlength{\tabcolsep}{3.55pt}
		\begin{tabular}[t]{lccc}
 		 \toprule
  		 Adversarial & BPD $\downarrow$     & IS $\uparrow$ & FID $\downarrow$\\
			\midrule 
		 mmd-GAN\cite{mmdgan} &       & $7.3$ & $25.0$  \\
		 SNGan~\cite{sngan} &        & $7.4$    & $29.3$ \\
						BatchGAN~\cite{lucas18icml} &        & $7.5$    & $23.7$ \\
			WGAN-GP~\cite{gulrajani17nips} &        & $7.9$  &  \\
			SNGAN$_{(R,H)}$ &       & {\bf 8.2}    & {\bf 21.7}\\
		   	\bottomrule 
		\end{tabular}
		\label{tab:sota_adv}
	} 
}
\newcommand{\MleTable}{
	{
		\setlength{\tabcolsep}{3pt}
		{\renewcommand{\arraystretch}{1.0}
		\begin{tabular}[t]{lccc}
		  \toprule
		   MLE & BPD $\downarrow$     & IS $\uparrow$ & FID $\downarrow$\\
			\midrule 
		   Real-NVP~\citep{realnvp}                  & $3.5$     & $4.5^{\dagger}$  & $56.8^{\dagger}$    \\ 						VAE-IAF~\citep{iaf}                      & 3.1    & $3.8^{\dagger}$        & $73.5^{\dagger}$    \\
						Pixcnn++ \citep{pixcnnpp}         & {\bf 2.9}    & 5.5    &    \\
			Flow++~\citep{Flow++}               & 3.1    &     &    \\
			Glow~\citep{Glow}          & 3.4 & {\bf 5.5}$^{\ddagger}$ & {\bf 46.8}$^{\ddagger}$ \\
		   														   	\bottomrule 
		\end{tabular}
		}
		\label{tab:sotaMlE}
	}
}

\newcommand{\stlTable}{
	{
		\setlength{\tabcolsep}{1pt}
		\scriptsize
		{\renewcommand{\arraystretch}{1.0}
		\begin{tabular}[t]{lccc}
		  \toprule
		  STL-10 ($48\times48$) & BPD $\downarrow$     & IS $\uparrow$ & FID $\downarrow$\\
			\midrule 
			\CQFG(wg, wd)         & $4.4$ & {\bf 9.4} & $44.3$ \\
			\CQFG(iaf, wd)         & $4.0$ & $8.6$ & $52.7$ \\
						\CQFG(s2)         & $3.8$ & $8.6$ & $52.1$ \\
			Real-NVP                  & {\bf 3.7}$^{\ddagger}$     & $4.8^{\ddagger}$  & $103.2^{\ddagger}$    \\ 		        			BatchGAN &        & 8.7 & 51\\
			SNGAN (Res-Hinge) &   & 9.1 & {\bf 40.1}\\
		   	\bottomrule 
		\end{tabular}
		}
	}
}

\newcommand{\ImagenetTable}{
	{
		\setlength{\tabcolsep}{1pt}
		\scriptsize
		{\renewcommand{\arraystretch}{1.2}
		\begin{tabular}[t]{lccc}
		  \toprule
		  ImageNet ($64\times64$) & BPD $\downarrow$     & IS $\uparrow$ & FID $\downarrow$\\
			\midrule 
			\CQFG(wg, wd)         & $4.90$ & $8.5$ & $45.5$ \\
			Real-NVP         & $3.98$ &  &  \\
			Glow         & $3.81$ &  &  \\
			Flow++         & {\bf 3.69} &  &  \\
			MMD-GAN         &  & {\bf 10.9} & {\bf 36.6} \\
								   	\bottomrule 
		\end{tabular}
		}
	}
}

\newcommand{\LsunTableMain}{
	{
		\setlength{\tabcolsep}{1pt}
		\scriptsize
		{\renewcommand{\arraystretch}{1.2}
			\begin{tabular}[t]{lccc}
						\toprule
						LSUN   	        & Real-NVP  & Glow & Ours\\ 
						\midrule
						Bedroom & ($2.72$/$\times$) & ($2.38$/$208.8^{\dagger}$) & ($3.91$, {\bf 21.1})\\
						Tower    & ($2.81$/$\times$) & ($2.46$/$214.5^{\dagger}$) & ($3.95$, {\bf 15.5})\\
						Church   & ($3.08$/$\times$) & ($2.67$/$222.3^{\dagger}$) & ($4.3$, {\bf 13.1}) \\
						Classroom & $\times$ & $\times$ & ($4.6$, $20.0$) \\
						Restaurant & $\times$ & $\times$ & ($4.7$, $20.5$) \\
																		\bottomrule
		\end{tabular}
	}
	}
}

\newcommand{\OtherDatasetTableOld}{
		\setlength{\tabcolsep}{3pt}
	\scriptsize
				\begin{tabular}{lccc}
						\toprule
						LSUN (BPD/FID)  	        & Real-NVP  & Glow & Ours\\ 
						\midrule
						LSUN-Bedrooms, 64 & (2.72/$\times$) & ({\bf 2.38}/\ble{$\times$}) & (3.91, 21.1)\\
						LSUN-Tower, 64    & (2.81/$\times$) & ({\bf 2.46}/\ble{$\times$}) & (3.95, 15.5)\\
						LSUN-Church, 64   & (3.08/$\times$) & ({\bf 2.67}/\ble{$\times$}) & (4.3, 13.1) \\
																		\bottomrule
		\end{tabular}						
					}

\newcommand{\cifarStlSamples}{
	{
																					\vspace*{2cm}
		\begin{tabular}[t]{c}
									\hspace*{-0.5cm} \setlength{\tabcolsep}{0pt}
\begin{tabular}{cccccccccc} 
\includegraphics{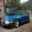} &
\includegraphics{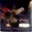} &
\includegraphics{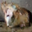} &
\includegraphics{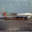} &
\includegraphics{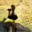} &
\includegraphics{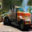} &
\includegraphics{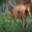} &
\includegraphics{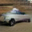} &
\includegraphics{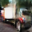} &
\includegraphics{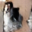}\\[-0.85ex]
\includegraphics{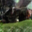} &
\includegraphics{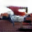} &
\includegraphics{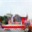} &
\includegraphics{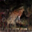} &
\includegraphics{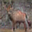} &
\includegraphics{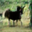} &
\includegraphics{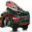} &
\includegraphics{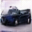} &
\includegraphics{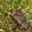} &
\includegraphics{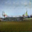}\\[-0.85ex]
\includegraphics{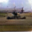} &
\includegraphics{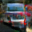} &
\includegraphics{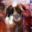} &
\includegraphics{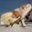} &
\includegraphics{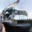} &
\includegraphics{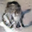} &
\includegraphics{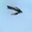} &
\includegraphics{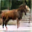} &
\includegraphics{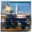} &
\includegraphics{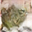}\\[-0.85ex]
\includegraphics{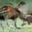} &
\includegraphics{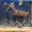} &
\includegraphics{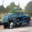} &
\includegraphics{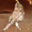} &
\includegraphics{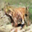} &
\includegraphics{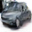} &
\includegraphics{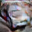} &
\includegraphics{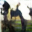} &
\includegraphics{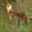} &
\includegraphics{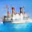}\\[-0.85ex]
\includegraphics{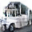} &
\includegraphics{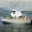} &
\includegraphics{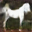} &
\includegraphics{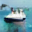} &
\includegraphics{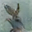} &
\includegraphics{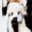} &
\includegraphics{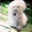} &
\includegraphics{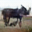} &
\includegraphics{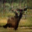} &
\includegraphics{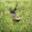}\\[-0.85ex]
\includegraphics{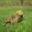} &
\includegraphics{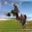} &
\includegraphics{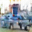} &
\includegraphics{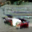} &
\includegraphics{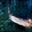} &
\includegraphics{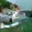} &
\includegraphics{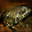} &
\includegraphics{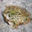} &
\includegraphics{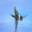} &
\includegraphics{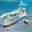}\\[-0.85ex]
\includegraphics{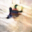} &
\includegraphics{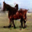} &
\includegraphics{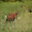} &
\includegraphics{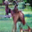} &
\includegraphics{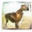} &
\includegraphics{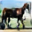} &
\includegraphics{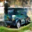} &
\includegraphics{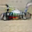} &
\includegraphics{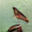} &
\includegraphics{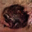}\\[-0.85ex]
\includegraphics{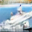} &
\includegraphics{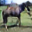} &
\includegraphics{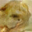} &
\includegraphics{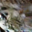} &
\includegraphics{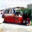} &
\includegraphics{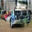} &
\includegraphics{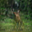} &
\includegraphics{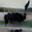} &
\includegraphics{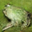} &
\includegraphics{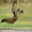}\\[-0.85ex]
\includegraphics{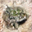} &
\includegraphics{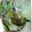} &
\includegraphics{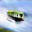} &
\includegraphics{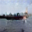} &
\includegraphics{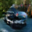} &
\includegraphics{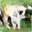} &
\includegraphics{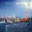} &
\includegraphics{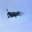} &
\includegraphics{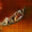} &
\includegraphics{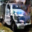}\\
\end{tabular} 
\\
						\hspace*{-0.5cm}  
		\end{tabular}
	}
		}

\newcommand{\sotaFig}{
	\begin{table}[tbh]
			\hybridTable
																						\end{table}
}

\newcommand{\refineTable}{
		\scriptsize 				\hspace*{-0.1cm}
		\tabcolsep=4pt
		\begin{tabular}{lccc}
			\toprule
			Refinements                     & BPD $\downarrow$ & IS $\uparrow$ & FID $\downarrow$\\			\midrule
			GAN                   & [$7.0$]            & $6.8$           & $31.4$  \\
			GAN (rd)             & [$6.9$]              & $7.4$           & $24.0$ \\
			\CQFG                  & $3.9$              & $7.1$           & $28.0$  \\
			\CQFG (rd)            & $3.8$              & $7.5$           & $26.0$   \\
						\CQFG (wg, rd)       & $3.8$                & {\bf 8.2}             & {\bf 17.2}\\
			\CQFG (iaf, rd)      & $3.7$            & $8.1$         & $18.6$   \\
			\CQFG (s2)            & {\bf 3.5}                & $6.9$             & $28.9$\\
			\bottomrule
		\end{tabular}
		\begin{center}
					\addtocounter{table}{1}
						Table \arabic{table}: Model refinements.
		\end{center}
		\label{tab:bells_whistles}
	}

\newcommand{\MLEandAdv}{
	\begin{figure}
		\hspace*{-1.0cm}
		\setlength{\tabcolsep}{1pt}
				\begin{tabular}{cc}\begin{tabular}{c} \setlength{\tabcolsep}{0pt} \includegraphics[width=0.51\textwidth]{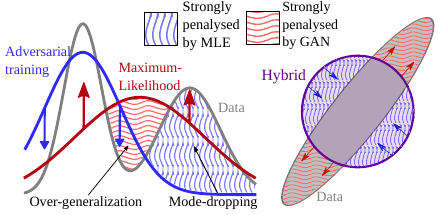}\end{tabular} & 
		\begin{minipage}[c]{0.45\linewidth}
		\addtocounter{figure}{1}
		\footnotesize
		Figure \arabic{figure}: (Left) Maximum likelihood training pulls probability mass towards high-density regions of the data distribution, while adversarial training pushes mass out of low-density regions. (Right) Independence assumptions become a source of conflict in a joint training setting, making hybrid training non-trivial. 		\end{minipage}
		\end{tabular}
		\vspace*{-0.7cm}
	\end{figure}
	\label{fig:cdt_qdt}
}

\newcommand{\LearnedSimilarity}{
									 		 		 		 				\begin{figure}
							\begin{minipage}[c]{0.47\textwidth}
												\includegraphics[width=0.99\textwidth]{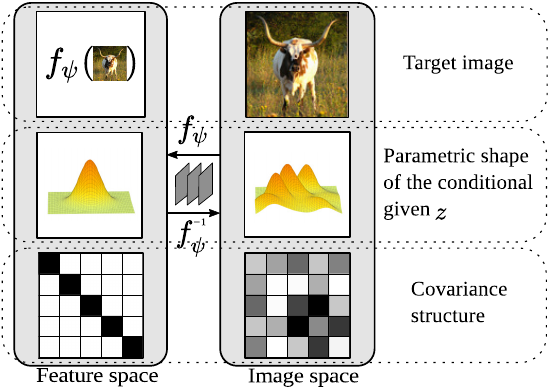}
				\addtocounter{figure}{1}
				\footnotesize
				Figure \arabic{figure}: 				An invertible non-linear mapping $f_{\psi}$ maps an image $x$ 
				to a vector $f_{\psi}(x)$ in feature space. 				$f_{\psi}$ is trained to adapt to modelling assumptions made by a trained 
				density $p_{\theta}$ in feature space.
				This induces full covariance structure and a non-Gaussian density. 
																																															\end{minipage}\hfill
			\begin{minipage}[c]{0.51\textwidth}
				\includegraphics[width=0.99\textwidth]{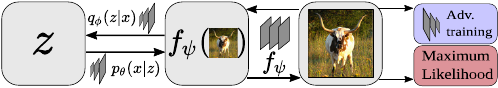}
								\addtocounter{figure}{1}
				\footnotesize
				Figure \arabic{figure}: 				Variational inference is used to train a latent variable generative model in feature space.
				The invertible mapping $f_{\psi}$ maps back to image space, where adversarial training can
				be performed together with MLE. \\
				\vspace*{-0.1cm}
				\begin{center}
				\setlength{\tabcolsep}{0pt}
				\begin{tabular}{cccc} 
				\includegraphics{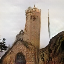} &
				\includegraphics{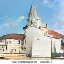} &
				\includegraphics{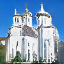} &
				\includegraphics{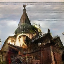} 
																												\end{tabular}
				\end{center}
				\addtocounter{figure}{1}
				\footnotesize
				Figure \arabic{figure}: 				Our model yields compelling samples while the optimization of likelihood ensures coverage of all modes in the training support and thus sample diversity, here on LSUN churches ($64 \times 64$). 			\end{minipage}
			\vspace*{-0.3cm}
	\end{figure}
		\label{fig:main_schema}
}

\newcommand{\highResImages}{
									\begin{tabular}{cc}
										\footnotesize
										Samples & Real images \\
					\toprule
						\hspace*{-0.3cm}  &  \hspace*{-0.3cm}  \\
					\midrule
					\hspace*{-0.3cm}  & \hspace*{-0.3cm}  \\
					\midrule
					\hspace*{-0.3cm}  & \hspace*{-0.3cm}  \\
					\midrule
					\hspace*{-0.3cm}  & \hspace*{-0.3cm}  \\
					\end{tabular}
				\figvspaceOne
							\caption{Samples and train images of  CelebA (crop 178), LSUN-Bedrooms, STL-10 96$\!\times\!$96, and ImageNet.}
								\label{fig:additionalSamples}			
								\figvspaceTwo
	}

\newcommand{\lsunHighRes}{
 

}

\newcommand{\lsunHighResRealBed}{
		\tabcolsep=2pt
		\begin{tabular}{cc}
			 & \setlength{\tabcolsep}{0pt}
\begin{tabular}{ccc} 
\includegraphics{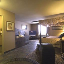} &
\includegraphics{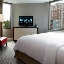} &
\includegraphics{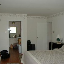}\\[-0.85ex]
\includegraphics{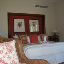} &
\includegraphics{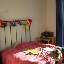} &
\includegraphics{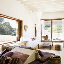}\\[-0.85ex]
\includegraphics{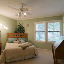} &
\includegraphics{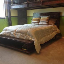} &
\includegraphics{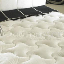}\\[-0.85ex]
\includegraphics{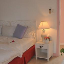} &
\includegraphics{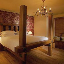} &
\includegraphics{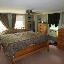}\\[-0.85ex]
\includegraphics{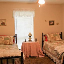} &
\includegraphics{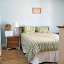} &
\includegraphics{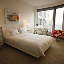}\\[-0.85ex]
\includegraphics{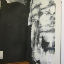} &
\includegraphics{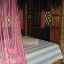} &
\includegraphics{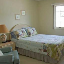}\\
\end{tabular} 

		\end{tabular}
	}
\newcommand{\lsunHighResRealTow}{
		\tabcolsep=2pt
		\begin{tabular}{cc}
			 & \setlength{\tabcolsep}{0pt}
\begin{tabular}{ccc} 
\includegraphics{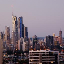} &
\includegraphics{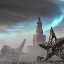} &
\includegraphics{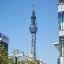}\\[-0.85ex]
\includegraphics{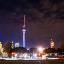} &
\includegraphics{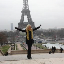} &
\includegraphics{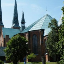}\\[-0.85ex]
\includegraphics{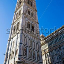} &
\includegraphics{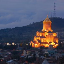} &
\includegraphics{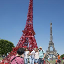}\\[-0.85ex]
\includegraphics{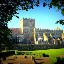} &
\includegraphics{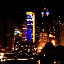} &
\includegraphics{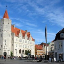}\\[-0.85ex]
\includegraphics{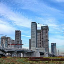} &
\includegraphics{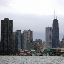} &
\includegraphics{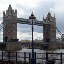}\\[-0.85ex]
\includegraphics{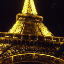} &
\includegraphics{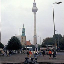} &
\includegraphics{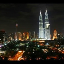}\\
\end{tabular} 
 \\
		\end{tabular}
}

\newcommand{\lsunHighResRealBri}{
		\tabcolsep=2pt
		\begin{tabular}{cc}
			 & \setlength{\tabcolsep}{0pt}
\begin{tabular}{ccc} 
\includegraphics{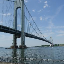} &
\includegraphics{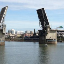} &
\includegraphics{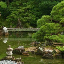}\\[-0.85ex]
\includegraphics{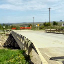} &
\includegraphics{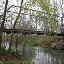} &
\includegraphics{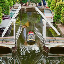}\\[-0.85ex]
\includegraphics{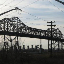} &
\includegraphics{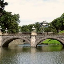} &
\includegraphics{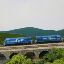}\\[-0.85ex]
\includegraphics{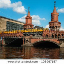} &
\includegraphics{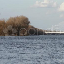} &
\includegraphics{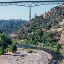}\\[-0.85ex]
\includegraphics{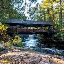} &
\includegraphics{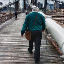} &
\includegraphics{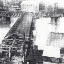}\\[-0.85ex]
\includegraphics{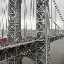} &
\includegraphics{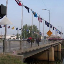} &
\includegraphics{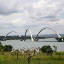}\\
\end{tabular} 
 \\
		\end{tabular}
}

\newcommand{\lsunHighResRealKit}{
		\tabcolsep=2pt
		\begin{tabular}{cc}
			 & \setlength{\tabcolsep}{0pt}
\begin{tabular}{ccc} 
\includegraphics{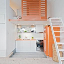} &
\includegraphics{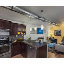} &
\includegraphics{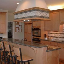}\\[-0.85ex]
\includegraphics{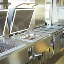} &
\includegraphics{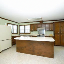} &
\includegraphics{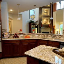}\\[-0.85ex]
\includegraphics{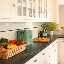} &
\includegraphics{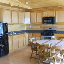} &
\includegraphics{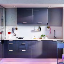}\\[-0.85ex]
\includegraphics{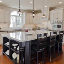} &
\includegraphics{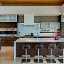} &
\includegraphics{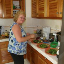}\\[-0.85ex]
\includegraphics{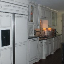} &
\includegraphics{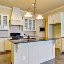} &
\includegraphics{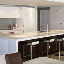}\\[-0.85ex]
\includegraphics{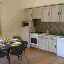} &
\includegraphics{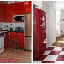} &
\includegraphics{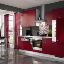}\\
\end{tabular} 

		\end{tabular}
}

\newcommand{\lsunHighResRealChu}{
		\tabcolsep=2pt
		\begin{tabular}{cc}
			 & \setlength{\tabcolsep}{0pt}
\begin{tabular}{ccc} 
\includegraphics{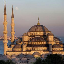} &
\includegraphics{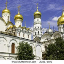} &
\includegraphics{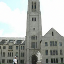}\\[-0.85ex]
\includegraphics{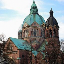} &
\includegraphics{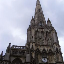} &
\includegraphics{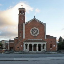}\\[-0.85ex]
\includegraphics{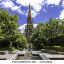} &
\includegraphics{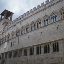} &
\includegraphics{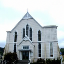}\\[-0.85ex]
\includegraphics{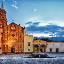} &
\includegraphics{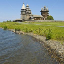} &
\includegraphics{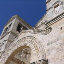}\\[-0.85ex]
\includegraphics{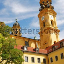} &
\includegraphics{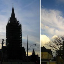} &
\includegraphics{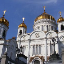}\\[-0.85ex]
\includegraphics{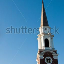} &
\includegraphics{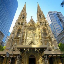} &
\includegraphics{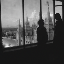}\\
\end{tabular} 

		\end{tabular}
}

\newcommand{\lsunHighResRealLiv}{
		\tabcolsep=2pt
		\begin{tabular}{cc}
			 & \setlength{\tabcolsep}{0pt}
\begin{tabular}{ccc} 
\includegraphics{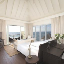} &
\includegraphics{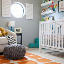} &
\includegraphics{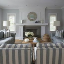}\\[-0.85ex]
\includegraphics{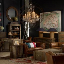} &
\includegraphics{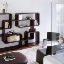} &
\includegraphics{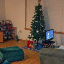}\\[-0.85ex]
\includegraphics{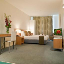} &
\includegraphics{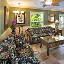} &
\includegraphics{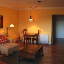}\\[-0.85ex]
\includegraphics{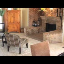} &
\includegraphics{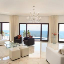} &
\includegraphics{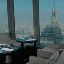}\\[-0.85ex]
\includegraphics{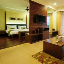} &
\includegraphics{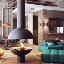} &
\includegraphics{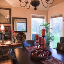}\\[-0.85ex]
\includegraphics{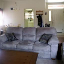} &
\includegraphics{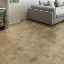} &
\includegraphics{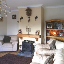}\\
\end{tabular} 

		\end{tabular}
}

\newcommand{\lsunHighResRealDin}{
		\tabcolsep=2pt
		\begin{tabular}{cc}
			 & \setlength{\tabcolsep}{0pt}
\begin{tabular}{ccc} 
\includegraphics{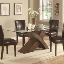} &
\includegraphics{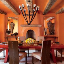} &
\includegraphics{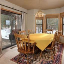}\\[-0.85ex]
\includegraphics{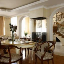} &
\includegraphics{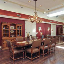} &
\includegraphics{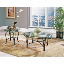}\\[-0.85ex]
\includegraphics{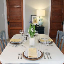} &
\includegraphics{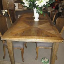} &
\includegraphics{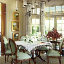}\\[-0.85ex]
\includegraphics{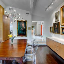} &
\includegraphics{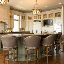} &
\includegraphics{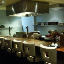}\\[-0.85ex]
\includegraphics{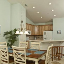} &
\includegraphics{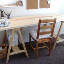} &
\includegraphics{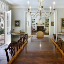}\\[-0.85ex]
\includegraphics{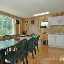} &
\includegraphics{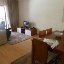} &
\includegraphics{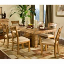}\\
\end{tabular} 

		\end{tabular}
}

\newcommand{\lsunHighResRealCon}{
		\tabcolsep=2pt
		\begin{tabular}{cc}
			 & \setlength{\tabcolsep}{0pt}
\begin{tabular}{ccc} 
\includegraphics{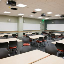} &
\includegraphics{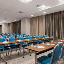} &
\includegraphics{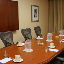}\\[-0.85ex]
\includegraphics{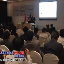} &
\includegraphics{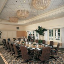} &
\includegraphics{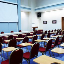}\\[-0.85ex]
\includegraphics{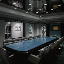} &
\includegraphics{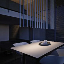} &
\includegraphics{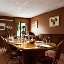}\\[-0.85ex]
\includegraphics{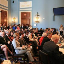} &
\includegraphics{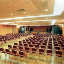} &
\includegraphics{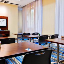}\\[-0.85ex]
\includegraphics{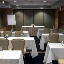} &
\includegraphics{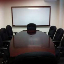} &
\includegraphics{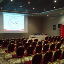}\\[-0.85ex]
\includegraphics{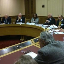} &
\includegraphics{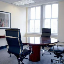} &
\includegraphics{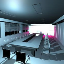}\\
\end{tabular} 

					\end{tabular}
}

\newcommand{\lsunHighResRealRest}{
		\tabcolsep=2pt
		\begin{tabular}{cc}
			 & \setlength{\tabcolsep}{0pt}
\begin{tabular}{ccc} 
\includegraphics{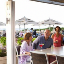} &
\includegraphics{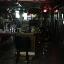} &
\includegraphics{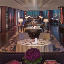}\\[-0.85ex]
\includegraphics{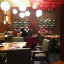} &
\includegraphics{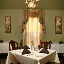} &
\includegraphics{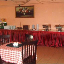}\\[-0.85ex]
\includegraphics{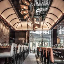} &
\includegraphics{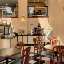} &
\includegraphics{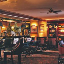}\\[-0.85ex]
\includegraphics{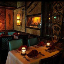} &
\includegraphics{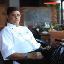} &
\includegraphics{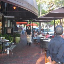}\\[-0.85ex]
\includegraphics{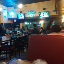} &
\includegraphics{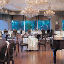} &
\includegraphics{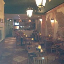}\\[-0.85ex]
\includegraphics{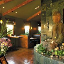} &
\includegraphics{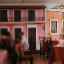} &
\includegraphics{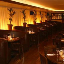}\\
\end{tabular} 

		\end{tabular}
}

\newcommand{\cifarAdditional}{
	\tabcolsep=2pt
	\hspace*{-0.4cm}
	\begin{tabular}{cc}
	 &
	\setlength{\tabcolsep}{0pt}
\begin{tabular}{ccccccc} 
\includegraphics{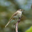} &
\includegraphics{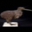} &
\includegraphics{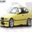} &
\includegraphics{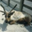} &
\includegraphics{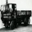} &
\includegraphics{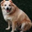} &
\includegraphics{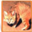}\\[-0.85ex]
\includegraphics{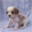} &
\includegraphics{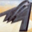} &
\includegraphics{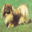} &
\includegraphics{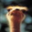} &
\includegraphics{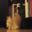} &
\includegraphics{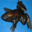} &
\includegraphics{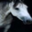}\\[-0.85ex]
\includegraphics{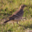} &
\includegraphics{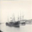} &
\includegraphics{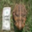} &
\includegraphics{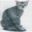} &
\includegraphics{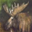} &
\includegraphics{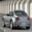} &
\includegraphics{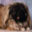}\\[-0.85ex]
\includegraphics{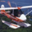} &
\includegraphics{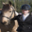} &
\includegraphics{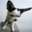} &
\includegraphics{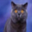} &
\includegraphics{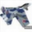} &
\includegraphics{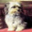} &
\includegraphics{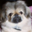}\\[-0.85ex]
\includegraphics{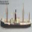} &
\includegraphics{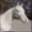} &
\includegraphics{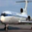} &
\includegraphics{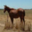} &
\includegraphics{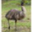} &
\includegraphics{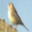} &
\includegraphics{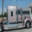}\\[-0.85ex]
\includegraphics{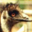} &
\includegraphics{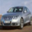} &
\includegraphics{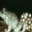} &
\includegraphics{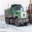} &
\includegraphics{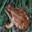} &
\includegraphics{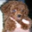} &
\includegraphics{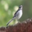}\\[-0.85ex]
\includegraphics{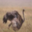} &
\includegraphics{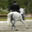} &
\includegraphics{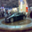} &
\includegraphics{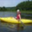} &
\includegraphics{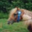} &
\includegraphics{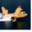} &
\includegraphics{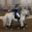}\\[-0.85ex]
\includegraphics{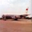} &
\includegraphics{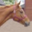} &
\includegraphics{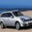} &
\includegraphics{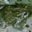} &
\includegraphics{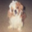} &
\includegraphics{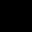} &
\includegraphics{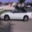}\\[-0.85ex]
\includegraphics{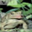} &
\includegraphics{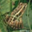} &
\includegraphics{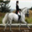} &
\includegraphics{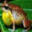} &
\includegraphics{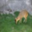} &
\includegraphics{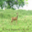} &
\includegraphics{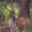}\\
\end{tabular} 
\\
	Samples from \CQFG (wg, rd) & Real images 
	\end{tabular}
}

\newcommand{\stlAdditional}{
	\tabcolsep=2pt
	\hspace*{-0.4cm}
	\begin{tabular}{cc}
		\setlength{\tabcolsep}{0pt}
\begin{tabular}{cccccccc} 
\includegraphics{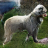} &
\includegraphics{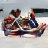} &
\includegraphics{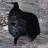} &
\includegraphics{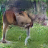} &
\includegraphics{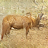} &
\includegraphics{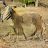} &
\includegraphics{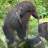} &
\includegraphics{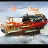}\\[-0.85ex]
\includegraphics{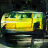} &
\includegraphics{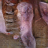} &
\includegraphics{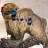} &
\includegraphics{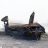} &
\includegraphics{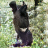} &
\includegraphics{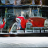} &
\includegraphics{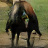} &
\includegraphics{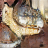}\\[-0.85ex]
\includegraphics{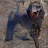} &
\includegraphics{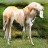} &
\includegraphics{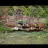} &
\includegraphics{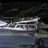} &
\includegraphics{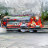} &
\includegraphics{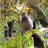} &
\includegraphics{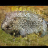} &
\includegraphics{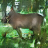}\\[-0.85ex]
\includegraphics{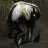} &
\includegraphics{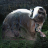} &
\includegraphics{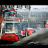} &
\includegraphics{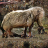} &
\includegraphics{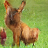} &
\includegraphics{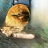} &
\includegraphics{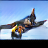} &
\includegraphics{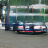}\\[-0.85ex]
\includegraphics{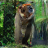} &
\includegraphics{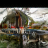} &
\includegraphics{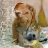} &
\includegraphics{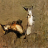} &
\includegraphics{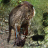} &
\includegraphics{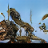} &
\includegraphics{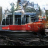} &
\includegraphics{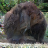}\\[-0.85ex]
\includegraphics{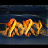} &
\includegraphics{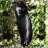} &
\includegraphics{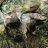} &
\includegraphics{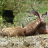} &
\includegraphics{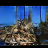} &
\includegraphics{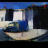} &
\includegraphics{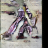} &
\includegraphics{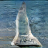}\\[-0.85ex]
\includegraphics{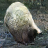} &
\includegraphics{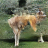} &
\includegraphics{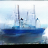} &
\includegraphics{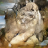} &
\includegraphics{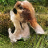} &
\includegraphics{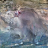} &
\includegraphics{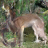} &
\includegraphics{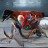}\\[-0.85ex]
\includegraphics{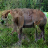} &
\includegraphics{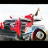} &
\includegraphics{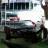} &
\includegraphics{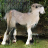} &
\includegraphics{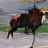} &
\includegraphics{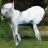} &
\includegraphics{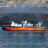} &
\includegraphics{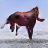}\\[-0.85ex]
\includegraphics{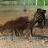} &
\includegraphics{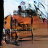} &
\includegraphics{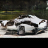} &
\includegraphics{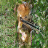} &
\includegraphics{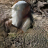} &
\includegraphics{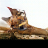} &
\includegraphics{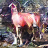} &
\includegraphics{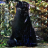}\\
\end{tabular} 
&
		\setlength{\tabcolsep}{0pt}
\begin{tabular}{cccc} 
\includegraphics{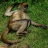} &
\includegraphics{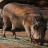} &
\includegraphics{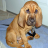} &
\includegraphics{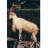}\\[-0.85ex]
\includegraphics{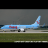} &
\includegraphics{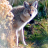} &
\includegraphics{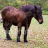} &
\includegraphics{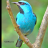}\\[-0.85ex]
\includegraphics{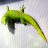} &
\includegraphics{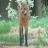} &
\includegraphics{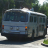} &
\includegraphics{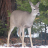}\\[-0.85ex]
\includegraphics{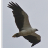} &
\includegraphics{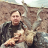} &
\includegraphics{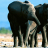} &
\includegraphics{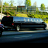}\\[-0.85ex]
\includegraphics{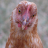} &
\includegraphics{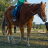} &
\includegraphics{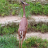} &
\includegraphics{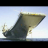}\\[-0.85ex]
\includegraphics{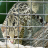} &
\includegraphics{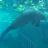} &
\includegraphics{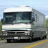} &
\includegraphics{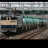}\\[-0.85ex]
\includegraphics{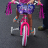} &
\includegraphics{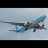} &
\includegraphics{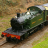} &
\includegraphics{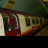}\\[-0.85ex]
\includegraphics{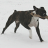} &
\includegraphics{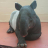} &
\includegraphics{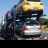} &
\includegraphics{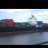}\\[-0.85ex]
\includegraphics{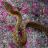} &
\includegraphics{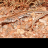} &
\includegraphics{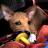} &
\includegraphics{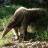}\\
\end{tabular} 
\\
		Samples from \CQFG (wg, rd) & Real images 
	\end{tabular}
}

\newcommand{\flowppComparisonCifar}{
	
}

\newcommand{\glowComparisonCifar}{
	\setlength{\tabcolsep}{0pt}
\begin{tabular}{ccccc} 
\includegraphics{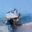} &
\includegraphics{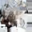} &
\includegraphics{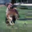} &
\includegraphics{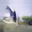} &
\includegraphics{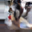}\\[-0.85ex]
\includegraphics{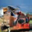} &
\includegraphics{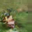} &
\includegraphics{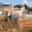} &
\includegraphics{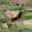} &
\includegraphics{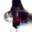}\\[-0.85ex]
\includegraphics{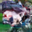} &
\includegraphics{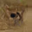} &
\includegraphics{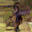} &
\includegraphics{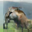} &
\includegraphics{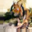}\\[-0.85ex]
\includegraphics{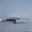} &
\includegraphics{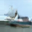} &
\includegraphics{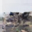} &
\includegraphics{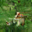} &
\includegraphics{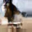}\\
\end{tabular} 

}

\newcommand{\nvpComparisonCifar}{
	
}

\newcommand{\flowGanAdvComparisonCifar}{
			\includegraphics[width=0.23\textwidth,trim={0 0 1.15cm 0},clip]{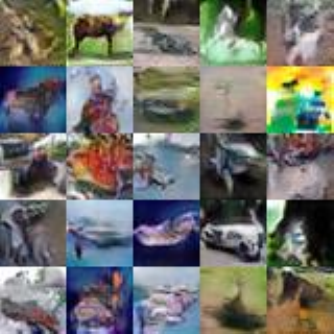}
}

\newcommand{\flowGanHybridComparisonCifar}{
			\includegraphics[width=0.23\textwidth,trim={0 0 1.15cm 0},clip]{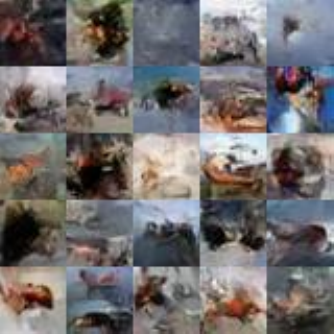}
}
\newcommand{\flowGanMLEComparisonCifar}{
			\includegraphics[width=0.23\textwidth,trim={0 0 1.15cm 0},clip]{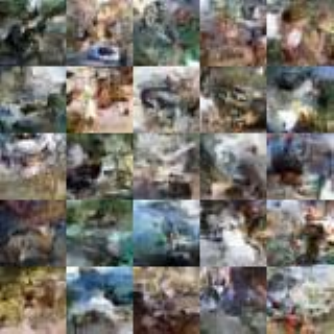}
}

\newcommand{\CifarComparisonFull}{
	\begin{tabular}{ccc}
		\hspace*{-0.15cm}  \hspace*{-0.9cm} &  \hspace*{-0.6cm} & \\
		flowpp & glow & nvp \\
				\includegraphics[width=0.29\textwidth,trim={0 2.30cm 0 0},clip]{cifar_gan5.pdf} & \includegraphics[width=0.29\textwidth,trim={0 2.30cm 0 0},clip]{cifar_hybrid5.pdf} & \includegraphics[width=0.29\textwidth,trim={0 2.30cm 0 0},clip]{cifar_mle5.pdf} \\
		Flowgan (Adversarial) & FlowGan (Hybrid) & FlowGan (MLE) \\
		\multicolumn{3}{c}{}
	\end{tabular}
}

\newcommand{\CifarComparison}{
	\setlength{\tabcolsep}{3pt}
	\begin{tabular}{ccc}
				 &
				\begin{tabular}{c}\includegraphics[width=0.29\textwidth,trim={0 1.15cm 0 0},clip]{cifar_hybrid5.pdf}\end{tabular} &
		\setlength{\tabcolsep}{0pt}
\begin{tabular}{cccccc} 
\includegraphics{./images/cifar_additional/cifar10/img0.png} &
\includegraphics{./images/cifar_additional/cifar10/img3.png} &
\includegraphics{./images/cifar_additional/cifar10/img6.png} &
\includegraphics{./images/cifar_additional/cifar10/img7.png} &
\includegraphics{./images/cifar_additional/cifar10/img8.png} &
\includegraphics{./images/cifar_additional/cifar10/img9.png}\\[-0.85ex]
\includegraphics{./images/cifar_additional/cifar10/img11.png} &
\includegraphics{./images/cifar_additional/cifar10/img12.png} &
\includegraphics{./images/cifar_additional/cifar10/img14.png} &
\includegraphics{./images/cifar_additional/cifar10/img16.png} &
\includegraphics{./images/cifar_additional/cifar10/img19.png} &
\includegraphics{./images/cifar_additional/cifar10/img20.png}\\[-0.85ex]
\includegraphics{./images/cifar_additional/cifar10/img21.png} &
\includegraphics{./images/cifar_additional/cifar10/img26.png} &
\includegraphics{./images/cifar_additional/cifar10/img30.png} &
\includegraphics{./images/cifar_additional/cifar10/img33.png} &
\includegraphics{./images/cifar_additional/cifar10/img34.png} &
\includegraphics{./images/cifar_additional/cifar10/img35.png}\\[-0.85ex]
\includegraphics{./images/cifar_additional/cifar10/img40.png} &
\includegraphics{./images/cifar_additional/cifar10/img41.png} &
\includegraphics{./images/cifar_additional/cifar10/img44.png} &
\includegraphics{./images/cifar_additional/cifar10/img46.png} &
\includegraphics{./images/cifar_additional/cifar10/img47.png} &
\includegraphics{./images/cifar_additional/cifar10/img48.png}\\
\end{tabular} 
 \\
		Glow @ $3.35$ BPD & FlowGan (H) @ $4.21$ BPD & \CQFG (iaf, rd) @ 3.7 BPD
	\end{tabular}
}

\newcommand{\LsunTable}{
		\footnotesize
		\tabcolsep=5pt
		\begin{tabular}{cc}
			\begin{tabular}{lcc}
			\toprule
				LSUN ($64 \times 64$) & BPD$\downarrow$ & FID$\downarrow$ \\
				\midrule
					Bedrooms   & 3.92 & 21.1 \\
					Towers     & 3.95 & 15.5 \\
					Bridges    & 4.1 & 25.2 \\
					Kitchen    & 4.0 & 15.3 \\
					Church     & 4.3 & 13.2 \\
					Living     & 4.1 & 16.7 \\
					Dining     & 4.3 & 14.6 \\
					Classroom  & 4.6 & 20.0 \\
					Conf. room & 4.2 & 21.3 \\
					Restaurant & 4.7 & 20.5 \\
				\bottomrule
				\end{tabular} &
			\begin{tabular}{lcc}
			\toprule
				LSUN ($128 \times 128$) & BPD$\downarrow$ & FID$\downarrow$ \\
				\midrule
				Bedrooms   & 3.0 &70.1 \\
				Towers     & 3.1 & 40.8 \\
				Bridges    & 3.4 & 64.2 \\
				Kitchen    & 3.1 & 45.2 \\
				Church     & 3.5 &47.3 \\
				Living     & 3.2 & 50.6 \\
				Dining     & 3.2 & 41.0\\
				Classroom  & 3.5 & 59.2 \\
				Conf. room & 3.1 &61.7 \\
				Restaurant & 3.8 & 60.0 \\
				\bottomrule
				\end{tabular} 
			\end{tabular}
}

\newcommand{\glowComparisonLsun}{
	\setlength{\tabcolsep}{0pt}
\begin{tabular}{ccc} 
\includegraphics{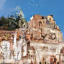} &
\includegraphics{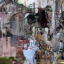} &
\includegraphics{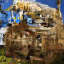}\\[-0.85ex]
\includegraphics{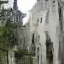} &
\includegraphics{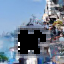} &
\includegraphics{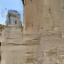}\\
\end{tabular} 
 \\
	\setlength{\tabcolsep}{0pt}
\begin{tabular}{ccc} 
\includegraphics{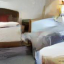} &
\includegraphics{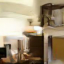} &
\includegraphics{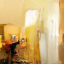}\\[-0.85ex]
\includegraphics{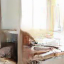} &
\includegraphics{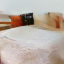} &
\includegraphics{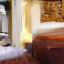}\\
\end{tabular} 
 \\
	
}

\newcommand{\nvpLsun}{
	 \\
	 \\
	 
}

\newcommand{\LsunComparison}{
	\tabcolsep=0.2pt
	\begin{tabular}{lcc}
																   &Glow~\citep{Glow} & Ours: \CQFG (wg, rd) \\
	{\footnotesize (C)} & & \setlength{\tabcolsep}{0pt}
\begin{tabular}{ccccc} 
\includegraphics{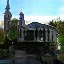} &
\includegraphics{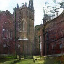} &
\includegraphics{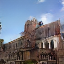} &
\includegraphics{./images/lsun64/lsun64_church_few/img9.png} &
\includegraphics{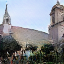}\\[-0.85ex]
\includegraphics{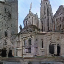} &
\includegraphics{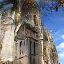} &
\includegraphics{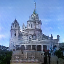} &
\includegraphics{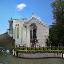} &
\includegraphics{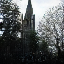}\\
\end{tabular} 
 \\
	{\footnotesize (B)} &  & \setlength{\tabcolsep}{0pt}
\begin{tabular}{ccccc} 
\includegraphics{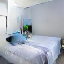} &
\includegraphics{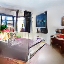} &
\includegraphics{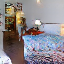} &
\includegraphics{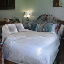} &
\includegraphics{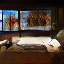}\\[-0.85ex]
\includegraphics{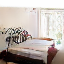} &
\includegraphics{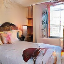} &
\includegraphics{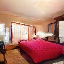} &
\includegraphics{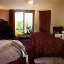} &
\includegraphics{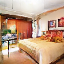}\\
\end{tabular} 

					\end{tabular}
}

\section{Introduction}
\newcommand\blfootnote[1]{  \begingroup
  \renewcommand\thefootnote{}\footnotetext{#1}    \endgroup
}
\blfootnote{${}^*$ The authors contributed equally.} 
\blfootnote{${}^{\dagger}$ Univ.\ Grenoble Alpes,   Inria, CNRS,  Grenoble INP, LJK,  38000 Grenoble, France.}
\blfootnote{${}^{\ddagger}$ Work done while at Inria.}

Successful recent generative models of natural images can be divided into two broad families, which are trained in fundamentally different ways.
The first is trained using likelihood-based criteria, which ensure that all training data points are well covered by the model.
This category includes variational autoencoders (VAEs)~\citep{iaf,vae,rezende15icml,rezende14icml}, autoregressive models such as PixelCNNs~\citep{pixcnnpp,pixrnn},
and flow-based models such as  Real-NVP~\citep{realnvp,Flow++,Glow}. 
The second category is trained based on a signal that measures to what extent (statistics of) samples from the model can be distinguished from (statistics of) the training data, \ie, based on the quality of samples drawn from the model.
This is the case for generative adversarial networks (GANs)~\citep{wgan,gan,karras18iclr}, and moment matching methods~\citep{mmd}.

Despite tremendous recent progress, existing methods exhibit a number of drawbacks.
Adversarially trained models such as GANs do not provide a density function, which poses a fundamental problem as it prevents assessment of how well the model fits held out and training data.
Moreover, adversarial models typically do not allow to infer the latent variables that underlie observed images. 
Finally, adversarial models suffer from mode collapse~\citep{wgan}, \ie, they do not cover the full support of the training data.
Likelihood-based model on the other hand are trained to put probability mass on all elements of the training set,
but over-generalise and produce samples of substantially inferior quality as compared to adversarial models.
The models with the best likelihood scores on held-out data are autoregressive models~\citep{highFidPixcnn}, which suffer from the additional problem that
they are extremely inefficient to sample from~\citep{fastpixcnn}, since images are generated pixel-by-pixel. The sampling inefficiency makes adversarial training of such models prohibitively expensive.

In order to overcome these shortcomings, we seek to design a model that (i) generates high-quality samples typical of adversarial models, (ii) provides a likelihood measure on the entire image space, and (iii) has a latent variable structure to enable efficient sampling and to permit adversarial training. Additionally we show that, (iv) a successful hybrid adversarial-likelihood paradigm requires going beyond simplifying assumptions commonly made with likelihood based latent variable models.
These simplifying assumptions on the conditional distribution on data $x$ given latents $z$,
$p(x|z)$, include full independence across the dimensions of $x$ and/or simple parametric forms such as Gaussian~\cite{vae}, or  use fully invertible networks~\cite{realnvp,Glow}.
These assumptions create a conflict between achieving high sample quality and high likelihood scores on held-out data.
Autoregressive models, such as pixelCNNs~\cite{pixrnn,pixcnnpp}, do not make factorization assumptions, but are extremely inefficient to sample from.
As a solution, we propose learning a non-linear invertible function $f_{\psi}$ between the image space and an abstract feature space, as illustrated in \fig{main_schema}.
Training a model with full support in this feature space induces a model in the image space that does not make Gaussianity, nor independence assumptions in the conditional density $p(x|z)$.
Trained by MLE, $f_{\psi}$ adapts to modelling assumptions made by $p_{\theta}$ so we refer to this approach as ''adaptive density estimation''.

\LearnedSimilarity

We  experimentally validate our approach on the CIFAR-10 dataset with an ablation study. Our model significantly improves over existing hybrid models, producing GAN-like samples, and IS and FID scores that are competitive with fully adversarial models, see Figure~3.
At the same time, we obtain likelihoods on held-out data comparable to state-of-the-art likelihood-based methods which requires covering the full support of the dataset.
We further confirm these observations with quantitative and qualitative experimental results on the STL-10, ImageNet  and LSUN datasets.

\section{Related work}
\label{sec:related}
Mode-collapse in GANs has  received considerable attention,  and stabilizing the training process as well as  improved and bigger architectures have been shown to alleviate this issue~\citep{wgan,gulrajani17nips,sngan}. 
Another line of work focuses on allowing the discriminator to access batch statistics of generated images, as pioneered by \citep{karras18iclr,salimans16nips}, 
and further generalized by \citep{pacgan,lucas18icml}. 
This enables comparison of distributional statistics by the discriminator rather than only  individual samples.
Other approaches to encourage diversity among  GAN samples include the use of maximum mean discrepancy~\cite{mmdgan}, optimal transport~\cite{otgan},
determinental point processes \cite{gdpp} and Bayesian formulations of adversarial training \cite{bayesianGAN} that allow model parameters to be sampled. 
In contrast to our work, these models lack an inference network, and do not define an explicit density over the full data support. 

An other line of research has explored inference mechanisms for GANs.  
 The discriminator of BiGAN~\citep{donahue17iclr} and ALI~\cite{dumoulin17iclr}, given pairs $(x,z)$ of images and latents, predict if $z$ was encoded from a real image, or if $x$ was decoded from a sampled $z$.
In \citep{ulyanov18aaai} the encoder and the discriminator are collapsed into one network that encodes both real images and generated samples, and tries to spread their posteriors apart.
In \citep{symVAE} a symmetrized KL divergence is approximated in an adversarial setup, and uses reconstruction losses to improve the correspondence between reconstructed and target variables for $x$ and $z$.
Similarly, in \citep{rosca2017variational} a discriminator is used to replace the KL divergence term in the variational lower bound used to train VAEs with the density ratio trick.
In \citep{makhzani16iclr} the KL divergence term in a VAE is replaced with a discriminator that compares latent variables from the prior and the posterior in a more flexible manner.
This regularization is more flexible than the standard KL divergence.
The VAE-GAN model~\citep{larsen16icml} and the model in \citep{featureConsistent} use the intermediate feature maps of a GAN discriminator and of a classifier respectively, as target space for a VAE.  
Unlike ours, these methods do not define a likelihood over the image space. 
\MLEandAdv

Likelihood-based models typically make modelling assumptions that conflict with adversarial training, these include strong factorization and/or Gaussianity. 
In our work we avoid these limitations by learning the shape of the conditional density on observed data given latents, $p(x|z)$, beyond fully factorized Gaussian models.
As in our work, Flow-GAN~\citep{flowgan} also builds on invertible transformations to construct a model that can be trained in a hybrid adversarial-MLE manner, see Figure 2.However, Flow-GAN does not use efficient non-invertible layers we introduce, and instead relies entirely on invertible layers.
	Other approaches combine autoregressive decoders with latent variable models to go beyond typical parametric assumptions in pixel space~\cite{chen17iclr,gulrajani17iclr,lucas18ecml}.
	They, however, are not amenable to adversarial training due to the prohibitively slow sequential pixel sampling.

\section{Preliminaries on MLE and adversarial training}
\label{sec:model}

\mypar{Maximum-likelihood and over-generalization} 
The de-facto standard approach to train generative models is maximum-likelihood estimation.
It maximizes the probability of data sampled from an unknown data distribution $\p$ under the model $\q$ \wrt the model parameters $\theta$.
This is equivalent to minimizing the Kullback-Leibler (KL) divergence,
$\kld{\p}{\q}$, between $\p$ and $\q$. This yields models that
tend to cover all the modes of the data, but put mass in
spurious regions of the target space; a phenomenon known
as ``over-generalization'' or ``zero-avoiding''~\citep{bishop06patrec}, 
and manifested by unrealistic samples in the context of generative image models, see Figure 4.Over-generalization is inherent to the optimization of the KL divergence oriented in this manner. 
Real images are sampled from $\p$, and $\q$ is explicitly optimized to cover all of them.
The training procedure, however, does not sample from $\q$ to evaluate the quality of such samples, ideally using the inaccessible $\p(x)$ as a score.
Therefore $\q$ may put mass in spurious regions of the space without being heavily penalized. 
We refer to this kind of training procedure as ``coverage-driven training'' (CDT).
This optimizes a loss of the form 
$\mathcal{L}_\text{C}(\q) = \int_{x} \p(x) \SC(x,\q) \;\text{d}x$,
where $\SC(x, \q) = \ln \q(x)$ evaluates how well a sample $x$ is covered by the model.
Any score $\SC$ that verifies: $\mathcal{L}_\text{C}(\q) = 0 \iff \q = \p$,
is equivalent to the log-score, which forms a justification for MLE on which we focus.

Explicitly evaluating sample quality is redundant in the regime of unlimited model capacity and  training data. 
Indeed, putting mass on spurious regions takes it away from the support of $\p$, and thus reduces the likelihood of the training data.
In practice, however, datasets and model capacity are finite, and 
models \emph{must} put mass outside the finite training set in order to generalize. 
The maximum likelihood criterion, by construction, only measures \emph{how much} mass goes off the training data, not \emph{where} it goes. 
In classic MLE, generalization is controlled in two ways: (i) inductive bias, in the form of model architecture, controls \emph{where} the off-dataset mass goes, and (ii) regularization controls to which extent this happens.
An adversarial loss, by considering samples of the model $\q$, can provide a second handle to evaluate and control where the off-dataset mass goes.
In this sense, and in contrast to model architecture design, an adversarial loss provides a ``trainable'' form of inductive bias.

\mypar{Adversarial models and mode collapse}
\label{sec:AdvModels}
Adversarially trained models produce samples of excellent quality.
As mentioned, their main drawbacks are their tendency to ``mode-drop'', and 
the lack of measure to assess mode-dropping, or their performance in general.
The reasons for this are two-fold. 
First, defining a valid likelihood requires adding volume to
the low-dimensional manifold learned by GANs to define a density under which training and test data have non-zero density.
Second, computing the density of a data point under the defined probability distribution requires marginalizing out the latent variables, which is not trivial in the absence of an efficient inference mechanism. 
When a human expert subjectively evaluates the quality of generated images, 
samples from the model are compared to the expert's implicit approximation of $\p$.
This type of objective may be formalized as $\mathcal{L}_\text{Q}(\q) = \int_{x} \q(x)s_q(x, \p)\; \textrm{d}x$, 
and we refer to it as ``quality-driven training'' (QDT).
To see that GANs \cite{gan} use this type of training, recall that 
the discriminator is trained with the loss
    $\mathcal{L}_{\text{GAN}} = \int_{x} \p(x) \ln D(x) + \q(x)\ln(1 - D(x)) \; \textrm{d} x.$
 It is easy to show that the optimal discriminator equals $ D^*(x) = \p(x) / (\p(x) + \q(x)).$
Substituting the optimal discriminator, $\mathcal{L}_{\text{GAN}}$ equals the Jensen-Shannon divergence, 
\begin{equation}
\JSD(\p || \q)= \frac{1}{2}\kld{\p}{\frac{1}{2}(\q+\p)} + \frac{1}{2}\kld{\q}{\frac{1}{2}(\q+\p)},
\end{equation}
up to additive and multiplicative constants~\citep{gan}.
This loss, approximated by the discriminator, is symmetric and  contains two KL divergence terms. 
Note that $\kld{\p}{\frac{1}{2}(\q+\p)}$ is an integral on $\p$, so \emph{coverage driven}.
The term that approximates it in $\mathcal{L}_{\text{GAN}}$, \ie, $\int_{x} \p(x) \ln D(x)$, is however independent 
from the generative model, and disappears when differentiating.
Therefore, it cannot be used to perform coverage-driven training, and 
the generator is trained to minimize $\ln(1 - D(G(z)))$ (or to maximize $\ln D(G(z))$),
where $G(z)$ is the deterministic generator that maps latent variables $z$ to the data space.
Assuming $D = D^*$, this yields
    					    \begin{equation}
	\int_z p(z)\ln(1-D^*(G(z))) \;\text{d}z = \int_x \q(x) \ln\frac{\q(x)}{\q(x) + \p(x)} \;\text{d}x = \kld{\q}{(\q + \p)/2},
\end{equation}
which is a quality-driven criterion, favoring sample quality over support coverage.

\vspace*{-0.1cm}
\section{Adaptive Density Estimation and hybrid adversarial-likelihood training}
\label{sec:ADE}

We present a hybrid training approach with MLE to cover the full support of the training data, and adversarial training as a trainable inductive bias mechanism to improve sample quality. Using both these criteria provides a richer training signal, but satisfying both criteria is more challenging than each in isolation for a given model complexity.  In practice, model flexibility is limited by (i) the number of parameters, layers, and features in the model, and (ii) simplifying modeling assumptions, usually made for tractability.
We show that these simplifying assumptions create a conflict between the two criteria, making successfull joint training non trivial. We introduce Adaptive Density Estimation as a solution to reconcile them. 
          
Latent variable generative models, defined as $\q(x) = \int_z \q(x|z)p(z)\; \text{d}z$, 
typically make simplifying assumptions on $\q(x|z)$, such as full factorization and/or Gaussianity, see \eg~\cite{dorta18cvpr,vae,litany18cvpr}. In particular, assuming full factorization of $\q(x|z)$ implies that 
any correlations not captured by $z$ are treated as independent per-pixel noise.
This is a poor model for natural images, unless $z$ captures each and every aspect of the image structure.
Crucially, this hypothesis is problematic in the context of hybrid MLE-adversarial training.
If $\p$ is too complex for $p_\theta(x|z)$ to fit it accurately enough, MLE will lead to a high variance in a factored (Gaussian) $p_\theta(x|z)$ as illustrated in Figure 4 (right).
This leads to unrealistic blurry samples, easily detected by an adversarial discriminator, which then does not provide a useful training signal.
Conversely, adversarial training will try to avoid these poor samples by dropping modes of the training data, and driving the ``noise'' level to zero.
This in turn is heavily penalized by maximum likelihood training, and leads to poor likelihoods on held-out data.
\label{sec:conflict}

\mypar{Adaptive density estimation} The point of view of regression hints at a possible solution. For instance, with isotropic Gaussian model densities with fixed variance, solving the optimization problem $ \theta^* \in \max_{\theta} \ln(p_{\theta}(x|z))$ is similar to solving 
$\min_{\theta} ||\mu_{\theta}(z) - x||_2$, i.e., $\ell_2$ regression, where $\mu_{\theta}(z)$ is the mean of the decoder $p_\theta(x|z)$. 
The Euclidean distance in RGB space is known to be a poor measure of similarity between images, non-robust to small translations or other basic transformations~\cite{mathieu16iclr}.
One can instead compute the Euclidean distance in a feature space, $|| f_{\psi}(x_1) - f_{\psi}(x_2)||_2 $,
where $f_{\psi}$ is chosen so that the distance is a better measure of similarity.
A popular way to obtain $f_{\psi}$ is to use a CNN that learns a non-linear image representation, that allows linear assessment of image similarity.
This is the idea underlying GAN discriminators, the FID evaluation measure~\citep{fid}, the reconstruction losses of VAE-GAN~\cite{larsen16icml} and classifier based perceputal losses as in \citep{featureConsistent}.

Despite their flexibility, such similarity metrics are in general degenerate in the sense that they may discard information about the data point $x$.
For instance, two different images $x$ and $y$ can collapse to the same points in feature space, i.e., $f_{\psi}(x) = f_{\psi}(y)$. 
This limits the use of similarity metrics in the context of generative modeling for two reasons: 
(i) it does not yield a valid measure of likelihood over inputs, and
(ii) points generated in the feature space $f_{\psi}$ cannot easily be mapped to images.
To resolve this issue, we chose $f_{\psi}$ to be a bijection.
Given a model $p_{\theta}$ trained to model $f_{\psi}(x)$ in feature space,
a density in image space is computed using the change of variable formula, which yields 
$ p_{\theta, \psi}(x) = p_{\theta}(f_{\psi}(x))\left|\textrm{det}\left(\partial f_{\psi}(x)/\partial x ^\top\right)\right|. $
\label{eq:COV} 
Image samples are obtained by sampling from $p_{\theta}$ in feature space, and mapping to the image space through $f_{\psi}^{-1}$.
We refer to this construction as Adaptive Denstiy Estimation.
If $p_{\theta}$ provides efficient log-likelihood computations, the change of variable formula can be used to train $f_{\psi}$ and $p_{\theta}$ together by maximum-likelihood, 
and if $p_{\theta}$ provides fast sampling adversarial training can be performed efficiently. 

\mypar{MLE with adaptive density estimation}
\label{sec:MLE_ADE}
To train a generative latent variable  model $\q(x)$ which permits efficient sampling, we rely on amortized variational inference.
We use an inference network $\enc (z | x)$ to construct a variational evidence lower-bound (ELBO),
						   \label{eq:elbo}
\begin{equation}
		\mathcal{L}_\text{ELBO}^{\psi}(x, \theta, \phi) = \mean_{ \enc (z | x)} \left[ \ln(\q(f_{\psi}(x) | z)) \right] - \kld{\enc (z|x)}{\q(z)} \vphantom{\log(\q(x | z))}  \leq \ln \q(f_{\psi}(x)).
\end{equation}
Using this lower bound together with the change of variable formula, the mapping to the similarity space $f_{\psi}$ and the generative model $p_{\theta}$ can be trained jointly with the loss  
	\begin{equation}
	\mathcal{L}_\text{C}(\theta, \phi, \psi)  =  \mean_{x \sim \p} \left[-\mathcal{L}_\text{ELBO}^{\psi}(x, \theta, \phi) - \ln\left|\det\DFDX\right|\right] \geq   -\mean_{x \sim \p} \left[\ln p_{\theta,\psi}(x)\right]. 
\end{equation}
We use gradient descent to train $f_{\psi}$ by optimizing $\mathcal{L}_\text{C}(\theta, \phi, \psi)$ \wrt $\psi$. 
The $\mathcal{L}_\text{ELBO}$ term encourges the mapping $f_{\psi}$ to maximize the density of points in feature space under the model $\q$, so that $f_{\psi}$ is trained to match modeling assumptions made in $\q$. 
Simultaneously, the log-determinant term
encourages $f_{\psi}$ to maximize the volume of data points in feature space. 
This guarantees that data points cannot be collapsed to a single point in the feature space.
We use a factored Gaussian form of the conditional $p_{\theta}(.|z)$ for tractability, but
since $f_{\psi}$ can arbitrarily reshape the corresponding conditional image space, it still avoids simplifying assumptions in the image space.
Therefore, the (invertible) transformation  $f_{\psi}$ avoids the conflict between the MLE and adversarial training mechanisms, and can leverage both.

\mypar{Adversarial training with adaptive density estimation} 
\label{sec:ADV_ADE}
To sample the generative model, we sample latents from the prior, $z\sim p_{\theta}(z)$,
which are then mapped to feature space through $\mu_\theta(z)$, and to image space through $f_{\psi}^{-1}$. 
We train our generator using the modified objective proposed by \citep{amap}, combining both generator losses considered in \citep{gan}, \ie $\ln[(1 - D(G(z)))/D(G(z))]$, which yields:
  \begin{equation}
	  \mathcal{L}_\text{Q}(p_{\theta,\psi}) = -\mean_{\q(z)}\left[ \ln D(f_{\psi}^{-1}(\mu_\theta(z))) - \ln( 1 - D(f_{\psi}^{-1}(\mu_\theta(z))))\right].
	\label{eq:LQ}
  \end{equation}
Assuming the discriminator $D$ is trained to optimality at every step, it is easy to demonstrate that the generator is trained to
optimize $\kld{p_{\theta, \psi}}{\p}$.
The training procedure, written as an algorithm in \app{algorithm}, alternates between 
(i) bringing $\mathcal{L}_\text{Q}(p_{\theta,\psi})$ closer to it's optimal value $\mathcal{L}_\text{Q}^*(p_{\theta,\psi}) = \kld{p_{\theta,\psi}}{\p}$, and
(ii) minimizing $\mathcal{L}_\text{C}(p_{\theta,\psi}) + \mathcal{L}_\text{Q}(p_{\theta,\psi})$. 
Assuming that the discriminator is trained to optimality at every step, the generative model is trained to minimize a bound on the symmetric sum of two KL divergences:
      $\mathcal{L}_\text{C}(p_{\theta,\psi}) + \mathcal{L}_\text{Q}^*(p_{\theta,\psi}) \geq  \kld{\p}{p_{\theta,\psi}} +  \kld{p_{\theta,\psi}}{\p} + \mathcal{H}(p^*),$
\label{eq:Full}
where the entropy of the data generating distribution,  $\mathcal{H}(p^*)$, is an additive constant independent of the generative model $p_{\theta,\psi}$. In contrast,
MLE and GANs optimize one of these divergences each.

\vspace*{-0.1cm}
\section{Experimental evaluation}
\vspace*{-0.1cm}
\label{sec:experiments}
We  present our evaluation protocol, followed by an ablation study to assess the importance of the  components of our model (\sect{ablation}).  We then show the quantitative and qualitative performance of our model,
and compare it to the state of the art on the CIFAR-10 dataset in \sect{sota}. 
We present additional results and comparisons on higher resolution datasets in \sect{datasets}. 

\label{sec:evaluation}
\mypar{Evalutation metrics} 
We evaluate our models with complementary metrics.
To assess sample quality, we report  the Fr\'echet inception distance (FID)~\citep{fid} and the inception score (IS)~\citep{salimans16nips}, which are the de facto standard metrics to evaluate GANs~\citep{sagan,biggan}. 
Although these metrics focus on sample quality, they are also sensitive to coverage, see \app{AppEval} for details.
To specifically evaluate the coverage of held-out data, we use the standard bits per dimension (BPD) metric, defined as the negative log-likelihood on held-out data, averaged across pixels and color channels~\citep{realnvp}.

Due to their degenerate low-dimensional support, GANs do not define a density in the image space, which prevents measuring BPD on them.
To endow a GAN with a full support and a likelihood, we train an inference network ``around it'', while keeping the weights of the GAN generator fixed. We also train an isotropic noise parameter $\sigma$. 
For both GANs and VAEs, we use this inference network to compute a lower bound to approximate the likelihood, \ie, an upper bound on BPD.
We evaluate all metrics using held-out data not used during training, which improves over common practice in the GAN literature, where training data is often used for evaluation. 

\subsection{Ablation study and comparison to VAE and GAN baselines}
\label{sec:ablation}	

We conduct an ablation study on the CIFAR-10 dataset.\footnote{We use the standard split of 50k/10k train/test images of 32$\!\times\!$32 pixels.}
Our GAN baseline uses the non-residual architecture of SNGAN~\citep{sngan}, which is stable and quick to train, without spectral normalization.
The same convolutional architecture is kept to build a VAE baseline.\footnote{In the VAE model, some intermediate feature maps are treated as conditional latent variables, allowing for hierarchical top-down sampling (see \app{implementation}).
Experimentally, we find that similar top-down sampling is not effective for the GAN model.}
It produces the mean of a factorizing Gaussian distribution. To ensure a valid density model we add a trainable isotropic variance $\sigma$. We train the generator for coverage by optimizing \mbox{$\mathcal{L}_Q(\q)$}, for quality by optimizing \mbox{$\mathcal{L}_C(\q)$}, and for both by optimizing the sum \mbox{$\mathcal{L}_Q(\q) + \mathcal{L}_C(\q)$}.
The model using Variational inference with Adaptive Density Estimation (ADE) is refered to as \piVAE. The addition of adversarial training is denoted \CQFG, and hybrid training with a Gaussian decoder as \CQG.
The bijective function $f_{\psi}$, implemented as a small Real-NVP with $1$ scale, $3$ residual blocks, $2$ layers per block, increases the number of weights by approximately 1.4\%. We compensate for these additional parameters with a slight decrease in the width of the generator for fair comparison.\footnote{This is, however, too small to have  a significant impact on the experimental results.}
See \app{implementation} for details. 

	\begin{figure}
		\begin{minipage}[c]{0.59\textwidth}
				\begin{tabular}{ccc}
					\hspace*{-.5cm} \setlength{\tabcolsep}{0pt}
\begin{tabular}{ccc} 
\includegraphics{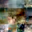} &
\includegraphics{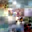} &
\includegraphics{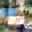}\\[-0.85ex]
\includegraphics{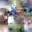} &
\includegraphics{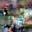} &
\includegraphics{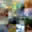}\\[-0.85ex]
\includegraphics{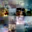} &
\includegraphics{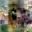} &
\includegraphics{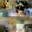}\\

\end{tabular} 
 & \hspace*{-0.20cm} \setlength{\tabcolsep}{0pt}
\begin{tabular}{ccc} 
\includegraphics{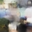} &
\includegraphics{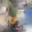} &
\includegraphics{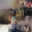}\\[-0.85ex]
\includegraphics{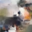} &
\includegraphics{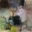} &
\includegraphics{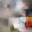}\\[-0.85ex]
\includegraphics{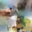} &
\includegraphics{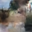} &
\includegraphics{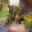}\\
\end{tabular} 
 & \hspace*{-0.20cm} \setlength{\tabcolsep}{0pt}
\begin{tabular}{ccc} 
\includegraphics{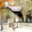} &
\includegraphics{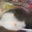} &
\includegraphics{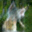}\\[-0.85ex]
\includegraphics{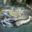} &
\includegraphics{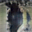} &
\includegraphics{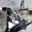}\\[-0.85ex]
\includegraphics{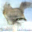} &
\includegraphics{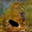} &
\includegraphics{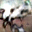}\\
\end{tabular} 
 \\
					VAE & \piVAE & \CQG \\
				\end{tabular} \\
				\begin{tabular}{cc}
					\hspace*{-0.5cm} \setlength{\tabcolsep}{0pt}
\begin{tabular}{ccccc} 
\includegraphics{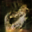} &
\includegraphics{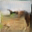} &
\includegraphics{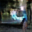} &
\includegraphics{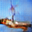} &
\includegraphics{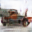}\\[-0.85ex]
\includegraphics{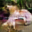} &
\includegraphics{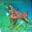} &
\includegraphics{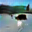} &
\includegraphics{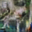} &
\includegraphics{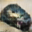}\\[-0.85ex]
\includegraphics{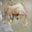} &
\includegraphics{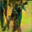} &
\includegraphics{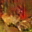} &
\includegraphics{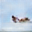} &
\includegraphics{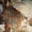}\\
\end{tabular} 
  &
					\hspace*{-0.5cm} \setlength{\tabcolsep}{0pt}
\begin{tabular}{ccccc} 
\includegraphics{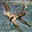} &
\includegraphics{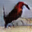} &
\includegraphics{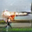} &
\includegraphics{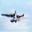} &
\includegraphics{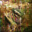}\\[-0.85ex]
\includegraphics{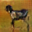} &
\includegraphics{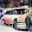} &
\includegraphics{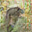} &
\includegraphics{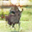} &
\includegraphics{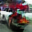}\\[-0.85ex]
\includegraphics{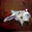} &
\includegraphics{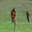} &
\includegraphics{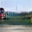} &
\includegraphics{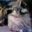} &
\includegraphics{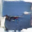}\\
\end{tabular} 
 \\
					 GAN & \CQFG (Ours)\\
				\end{tabular}
				\figvspaceOne 
		\end{minipage} \hfill
		\begin{minipage}[c]{0.39\textwidth}
			\setlength{\tabcolsep}{2pt}
			\footnotesize
			\begin{tabular}{lccccccr}
				\toprule
				& $f_{\psi}$ & Adv. & {\scriptsize MLE} & {\scriptsize BPD} $\downarrow$ & {\scriptsize IS} $\uparrow$ & {\scriptsize FID} $\downarrow$\\ 				\midrule
								{\scriptsize GAN}                                                                                        & \N & \Y           & \N            & [7.0]                & 6.8           & 31.4  \\
								{\scriptsize VAE}                                                                                        & \N & \N           & \Y            & 4.4              & 2.0           & 171.0 \\

				{\scriptsize \piVAE}$^{\dagger}$                                                                                       &\Y & \N           & \Y           & 3.5       & 3.0          & 112.0   \\
				\midrule
								{\scriptsize \CQG}  & \N & \victory           & \victory                   & 4.4    & 5.1  & 58.6 \\
								{\scriptsize \CQFG}$^{\dagger}$                                                                                        & \Y & \victory           & \victory           & 3.9              & 7.1           & 28.0   \\
				\bottomrule
			\end{tabular}
					
					\footnotesize
					\addtocounter{figure}{1}
					\addtocounter{table}{1}
					\vspace*{0.2cm}
					Table \arabic{table}: Quantitative results. $^{\dagger}$ : Parameter count decreased by $1.4\%$ to compensate for $f_{\psi}$.
					[Square brackets] denote that the value is approximated, see \sect{evaluation}.\\

					Figure \arabic{figure}: Samples from GAN and VAE baselines, our \piVAE,  \CQG and \CQFG models, all trained on CIFAR-10.
												\end{minipage}
	\end{figure}
	\def \tabBaselineCount {\arabic{table}}
	\def \figBaselineCount {\arabic{figure}}
		\label{fig:BaselineSamples}
	\label{tab:baselines}

Experimental results in Table 1 confirm that the GAN baseline yields better sample quality (IS and FID) than the VAE baseline: obtaining inception scores of $6.8$ and $2.0$, respectively.
Conversely, VAE achieves better coverage, with a BPD of $4.4$, compared to $7.0$ for GAN.
An identical generator trained for both quality and coverage, \CQG, obtains a sample quality that is in between that of the GAN and the VAE baselines, in line with the analysis in \sect{conflict}.
Samples from the different models in Figure~5 confirm these quantitative results. 
Using $f_{\psi}$ and training with $\mathcal{L}_\textrm{C}(\q)$ only, denoted by \piVAE in the table, leads to improved sample quality with IS up from $2.0$ to $3.0$ and FID down from $171$ to $112$. Note that this quality is still below the GAN baseline and our \CQG model.

When $f_{\psi}$ is used with coverage and quality driven training, \CQFG, we obtain improved IS and FID scores over the GAN baseline, with IS up from $6.8$ to $7.1$, and FID down from $31.4$ to $28.0$. The examples shown in the figure confirm the high quality of the samples generated by our \CQFG model. Our model also achieves a better BPD than the VAE baseline.
These experiments demonstrate that our proposed bijective feature space substantially improves the compatibility of coverage and quality driven training. We obtain improvements over both VAE and GAN in terms of held-out likelihood, and improve VAE sample quality to, or beyond, that of GAN. We further evaluate our model using the recent precision and recall approach of~\citep{prdEval} an the classification framework of~\citep{Ktest} in \app{prdKtrain}.
Additional results showing the impact of the number of layers and scales in the bijective similarity mapping $f_{\psi}$ (\app{features}), reconstructions qualitatively demonstrating the inference abilities of our \CQFG model (\app{Reconstruct}) are presented in the supplementary material.

\subsection{Refinements and comparison to the state of the art}
\label{sec:refinements}

\begin{minipage}[c]{0.69\textwidth}
We now consider further refinements to our model, inspired by recent generative modeling literature. Four refinements are used: (i) adding residual connections to
the discriminator~\citep{gulrajani17nips} (rd), (ii) leveraging more accurate posterior
approximations using inverse auto-regressive flow~\citep{iaf} (iaf); see
\app{implementation}, (iii) training wider generators with twice as many
channels (wg), and (iv) using a hierarchy of two scales to build $f_{\psi}$
(s2); see \app{features}. Table~\hyperref[tab:bells_whistles]{2} shows
consistent improvements with these additions, in terms of BPD, IS, FID.
\end{minipage}\hfill
\begin{minipage}[c]{0.29\textwidth}
	\hspace*{0cm}
	\refineTable
\end{minipage}
\label{sec:sota}\\

\begin{figure}[h]
\begin{minipage}[c]{0.69\textwidth}
Table \hyperref[tab:sota_cifar]{3} compares our model to existing hybrid approaches and state-of-the-art generative models on CIFAR-10.
In the category of hybrid models that define a valid likelihood over the data space, denoted by Hybrid (L) in the table,
FlowGAN(H) optimizes MLE and an adversarial loss, and FlowGAN(A) is trained adversarially.
The \CQFG model significantly outperforms these two variants both in terms of BPD, from $4.2$ to between $3.5$ and $3.8$, 
and quality, \eg, IS improves from $5.8$ to $8.2$.
Compared to models that train an inference network adversarially, denoted by Hybrid (A),
our model shows a substantial improvement in IS from $7.0$ to $8.2$. Note that these models 
do not allow likelihood evaluation, thus BPD values are not defined.\\

Compared to adversarial models, which are not optimized for support coverage,
\CQFG obtains better FID ($17.2$ down from $21.7$) and similar IS ($8.2$ for both)
compared to SNGAN with residual connections and hinge-loss,
despite training on $17 \%$ less data than GANs (test split removed). 
The improvement in FID is likely due to this measure being more sensitive to support coverage than IS.
Compared to models optimized with MLE only, we obtain a BPD between $3.5$ and $3.7$, comparable to $3.5$ for Real-NVP demonstrating a good coverage of the support of held-out data.
We computed IS and FID scores for MLE based models using publicly released code, with provided parameters (denoted by $^{\dagger}$ in the table) or trained ourselves (denoted by $^{\ddagger}$).
Despite being smaller (for reference Glow has $384$ layers \vs at most $10$ for our deeper generator), our \CQFG model generates better samples, \eg, IS up from $5.5$ to $8.2$ (samples displayed in \fig{cifcomp}), owing to quality driven training controling where the off-dataset mass goes.
Additional samples from our \CQFG model and comparison to others models are given in \app{samples}.

\end{minipage} \hfill
\begin{minipage}[l]{0.29\textwidth}
	        \vspace*{-0.5cm}
		\hspace*{-0.5cm}
		\begin{tabular}{l}
			\scriptsize \hybridTable \\
			\scriptsize \advHybridTable \\
			\scriptsize \AdvTable \\
			\scriptsize \MleTable \\
		\end{tabular}
		\addtocounter{table}{1}
				\scriptsize
		Table \arabic{table}: Performance on CIFAR10, without labels. MLE and Hybrid (L) models discard the test split. $\dagger$: computed by us using provided weights. $\ddagger$: computed by us using provided code to (re)train models.
		\label{tab:sota_cifar}
\end{minipage}
\end{figure}

\begin{figure}
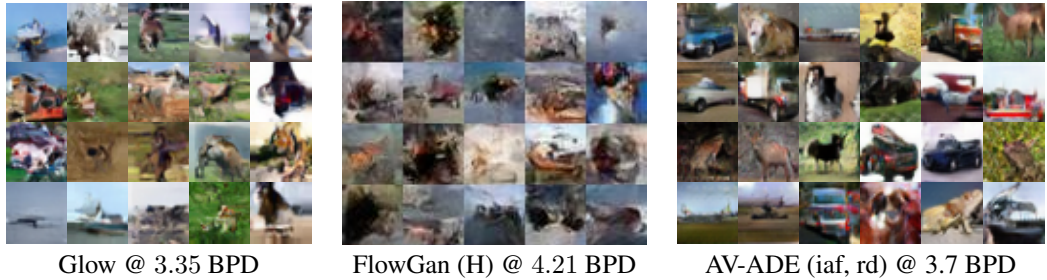

	
	\setlength{\tabcolsep}{3pt}
	\begin{tabular}{ccc}
				\setlength{\tabcolsep}{0pt}
\begin{tabular}{ccccc} 
\includegraphics{./images/glow_nvp/cifar/img0.png} &
\includegraphics{./images/glow_nvp/cifar/img1.png} &
\includegraphics{./images/glow_nvp/cifar/img2.png} &
\includegraphics{./images/glow_nvp/cifar/img3.png} &
\includegraphics{./images/glow_nvp/cifar/img4.png}\\[-0.85ex]
\includegraphics{./images/glow_nvp/cifar/img5.png} &
\includegraphics{./images/glow_nvp/cifar/img6.png} &
\includegraphics{./images/glow_nvp/cifar/img7.png} &
\includegraphics{./images/glow_nvp/cifar/img8.png} &
\includegraphics{./images/glow_nvp/cifar/img9.png}\\[-0.85ex]
\includegraphics{./images/glow_nvp/cifar/img10.png} &
\includegraphics{./images/glow_nvp/cifar/img11.png} &
\includegraphics{./images/glow_nvp/cifar/img12.png} &
\includegraphics{./images/glow_nvp/cifar/img13.png} &
\includegraphics{./images/glow_nvp/cifar/img14.png}\\[-0.85ex]
\includegraphics{./images/glow_nvp/cifar/img15.png} &
\includegraphics{./images/glow_nvp/cifar/img16.png} &
\includegraphics{./images/glow_nvp/cifar/img17.png} &
\includegraphics{./images/glow_nvp/cifar/img18.png} &
\includegraphics{./images/glow_nvp/cifar/img19.png}\\
\end{tabular} 
 &
				\begin{tabular}{c}\includegraphics[width=0.29\textwidth,trim={0 1.15cm 0 0},clip]{cifar_hybrid5.pdf}\end{tabular} &
		\setlength{\tabcolsep}{0pt}
\begin{tabular}{cccccc} 
\includegraphics{./images/cifar_additional/cifar10/img0.png} &
\includegraphics{./images/cifar_additional/cifar10/img3.png} &
\includegraphics{./images/cifar_additional/cifar10/img6.png} &
\includegraphics{./images/cifar_additional/cifar10/img7.png} &
\includegraphics{./images/cifar_additional/cifar10/img8.png} &
\includegraphics{./images/cifar_additional/cifar10/img9.png}\\[-0.85ex]
\includegraphics{./images/cifar_additional/cifar10/img11.png} &
\includegraphics{./images/cifar_additional/cifar10/img12.png} &
\includegraphics{./images/cifar_additional/cifar10/img14.png} &
\includegraphics{./images/cifar_additional/cifar10/img16.png} &
\includegraphics{./images/cifar_additional/cifar10/img19.png} &
\includegraphics{./images/cifar_additional/cifar10/img20.png}\\[-0.85ex]
\includegraphics{./images/cifar_additional/cifar10/img21.png} &
\includegraphics{./images/cifar_additional/cifar10/img26.png} &
\includegraphics{./images/cifar_additional/cifar10/img30.png} &
\includegraphics{./images/cifar_additional/cifar10/img33.png} &
\includegraphics{./images/cifar_additional/cifar10/img34.png} &
\includegraphics{./images/cifar_additional/cifar10/img35.png}\\[-0.85ex]
\includegraphics{./images/cifar_additional/cifar10/img40.png} &
\includegraphics{./images/cifar_additional/cifar10/img41.png} &
\includegraphics{./images/cifar_additional/cifar10/img44.png} &
\includegraphics{./images/cifar_additional/cifar10/img46.png} &
\includegraphics{./images/cifar_additional/cifar10/img47.png} &
\includegraphics{./images/cifar_additional/cifar10/img48.png}\\
\end{tabular} 
 \\
		Glow @ $3.35$ BPD & FlowGan (H) @ $4.21$ BPD & \CQFG (iaf, rd) @ 3.7 BPD
	\end{tabular}

	\caption{\footnotesize Samples from models trained on CIFAR-10. Our \CQFG spills less mass on unrealistic samples, owing to adversarial training which controls where off-dataset mass goes. \vspace*{-0.7cm}}
	\label{fig:cifcomp}
\end{figure}

\subsection{Results on additional datasets}
\label{sec:datasets}
To further validate our model we  evaluate it on STL10 ($48 \times 48$), ImageNet and LSUN (both $64 \times 64$).
We use a wide generator to account for the higher resolution, without IAF, a single scale in $f_{\psi}$, and no residual blocks
(see \sect{refinements}).
The architecture and training hyper-parameters are not tuned, besides adding one layer at resolution $64 \times 64$, which demonstrates the stability of our approach. On STL10, Table \hyperref[tab:sota_cifar]{4} shows that our \CQFG improves inception score over SNGAN, from $9.1$ up to $9.4$, and is second best in FID. Our likelihood performance, between $3.8$ and $4.4$, and close to that of Real-NVP at $3.7$, demonstrates good coverage of the support of held-out data. On the ImageNet dataset, maintaining high sample quality, while covering the full support is challenging, due to its very diverse support. Our \CQFG model obtains a sample quality behind that of MMD-GAN with IS/FID scores at $8.5/45.5$ \vs $10.9/36.6$. However, MMD-GAN is trained purely adversarially and does not provide a valid density across the data space, unlike our approach.

\fig{lsun_comp} shows samples from our generator trained on a single GPU with $11$ Gb of memory on LSUN classes. 
Our model yields more compelling samples compared to those of Glow, despite having less layers (7 \vs over 500).
Additional samples on other LSUN categories are presented in \app{samples}.

\vspace*{-0.5cm}
\begin{table}[htb]
	\hspace*{-0.3cm}
	\setlength{\tabcolsep}{0pt}
	\begin{tabular}{ccc}
		\begin{tabular}{c} \stlTable \end{tabular} \hspace*{-0.2cm} & \begin{tabular}{c} \ImagenetTable \end{tabular} \hspace*{-0.2cm} & \begin{tabular}{c}\LsunTableMain \end{tabular} 
	\end{tabular}
	\caption{\footnotesize Results on the STL-10, ImageNet, and LSUN datasets. \CQFG (wg, rd) is used for LSUN.}
	\label{tab:sota}
\end{table}

\begin{figure}[hbt]
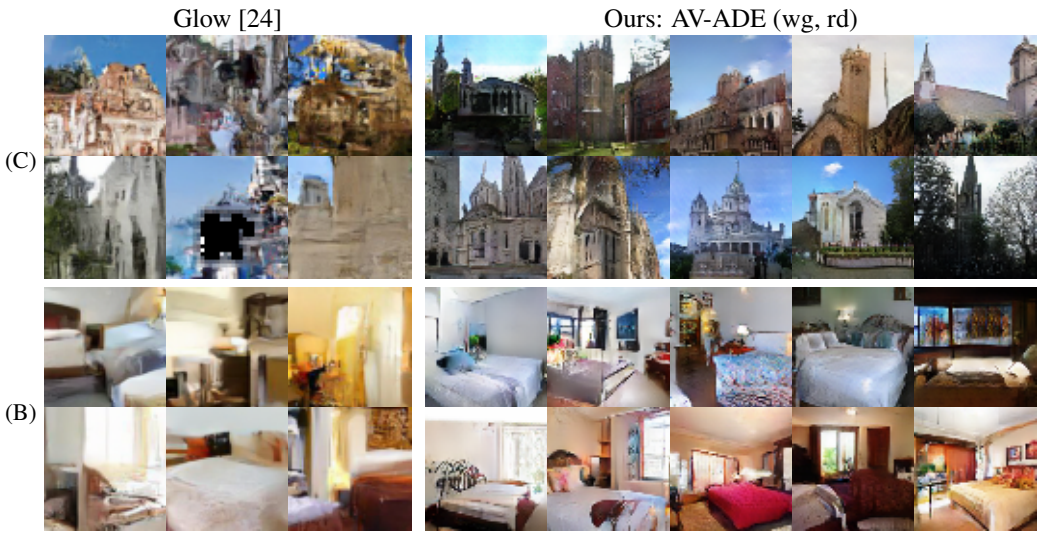

	\hspace*{-0.8cm}
	
	\tabcolsep=0.2pt
	\begin{tabular}{lcc}
																   &Glow~\citep{Glow} & Ours: \CQFG (wg, rd) \\
	{\footnotesize (C)} &\setlength{\tabcolsep}{0pt}
\begin{tabular}{ccc} 
\includegraphics{./images/glow_nvp/church/img0.png} &
\includegraphics{./images/glow_nvp/church/img1.png} &
\includegraphics{./images/glow_nvp/church/img3.png}\\[-0.85ex]
\includegraphics{./images/glow_nvp/church/img4.png} &
\includegraphics{./images/glow_nvp/church/img5.png} &
\includegraphics{./images/glow_nvp/church/img7.png}\\
\end{tabular} 
 & \setlength{\tabcolsep}{0pt}
\begin{tabular}{ccccc} 
\includegraphics{./images/lsun64/lsun64_church_few/img3.png} &
\includegraphics{./images/lsun64/lsun64_church_few/img4.png} &
\includegraphics{./images/lsun64/lsun64_church_few/img8.png} &
\includegraphics{./images/lsun64/lsun64_church_few/img9.png} &
\includegraphics{./images/lsun64/lsun64_church_few/img10.png}\\[-0.85ex]
\includegraphics{./images/lsun64/lsun64_church_few/img11.png} &
\includegraphics{./images/lsun64/lsun64_church_few/img13.png} &
\includegraphics{./images/lsun64/lsun64_church_few/img14.png} &
\includegraphics{./images/lsun64/lsun64_church_few/img19.png} &
\includegraphics{./images/lsun64/lsun64_church_few/img20.png}\\
\end{tabular} 
 \\
	{\footnotesize (B)} & \setlength{\tabcolsep}{0pt}
\begin{tabular}{ccc} 
\includegraphics{./images/glow_nvp/bedroom/img0.png} &
\includegraphics{./images/glow_nvp/bedroom/img2.png} &
\includegraphics{./images/glow_nvp/bedroom/img3.png}\\[-0.85ex]
\includegraphics{./images/glow_nvp/bedroom/img4.png} &
\includegraphics{./images/glow_nvp/bedroom/img5.png} &
\includegraphics{./images/glow_nvp/bedroom/img7.png}\\
\end{tabular} 
 & \setlength{\tabcolsep}{0pt}
\begin{tabular}{ccccc} 
\includegraphics{./images/lsun64/lsun64_bedroom_few/img1.png} &
\includegraphics{./images/lsun64/lsun64_bedroom_few/img2.png} &
\includegraphics{./images/lsun64/lsun64_bedroom_few/img13.png} &
\includegraphics{./images/lsun64/lsun64_bedroom_few/img16.png} &
\includegraphics{./images/lsun64/lsun64_bedroom_few/img20.png}\\[-0.85ex]
\includegraphics{./images/lsun64/lsun64_bedroom_few/img21.png} &
\includegraphics{./images/lsun64/lsun64_bedroom_few/img24.png} &
\includegraphics{./images/lsun64/lsun64_bedroom_few/img26.png} &
\includegraphics{./images/lsun64/lsun64_bedroom_few/img27.png} &
\includegraphics{./images/lsun64/lsun64_bedroom_few/img32.png}\\
\end{tabular} 

					\end{tabular}

	\hspace*{-0.4cm}
	\caption{\footnotesize Samples from models trained on LSUN Churches (C) and bedrooms (B). Our \CQFG model over-generalises less and produces more compelling samples. See \app{samples} for more classes and samples.}
	\label{fig:lsun_comp}
\end{figure}

\vspace*{-0.4cm}
\section{Conclusion}
\vspace*{-0.2cm}
We presented a generative model that leverages invertible network layers to relax the conditional independence assumptions commonly made in VAE decoders.
It allows for efficient feed-forward sampling, and can be trained using a maximum likelihood criterion that ensures coverage of the data generating distribution, as well as an adversarial criterion that ensures high  sample quality.

\mypar{Acknowledgments}
The authors would like to thank Corentin Tallec, Mathilde Caron, Adria Ruiz and Nikita Dvornik for useful feedback and discussions.
Acknowledgments also go to our anonymous reviewers, who contributed valuable comments and remarks.\\
This work was supported in part by the grants ANR16-CE23-0006 “Deep in France”, LabEx PERSYVAL-Lab (ANR-11-LABX0025-01) as well as the Indo-French project EVEREST (no. 5302-1) funded by CEFIPRA and a grant from ANR (AVENUE project ANR-18-CE23-0011),
\bibliographystyle{abbrvnat}
\bibliography{jjv,mllgan}
\appendix
\clearpage
\begin{section}{Quantitative and qualitative results}
		In this section we provide additional quantitative and qualitative results on CIFAR10, STL10,
	LSUN (categories: Bedrooms, Towers, Bridges, Kitchen, Church, Living room, Dining room, Classroom, Conference room and Restaurant) at resolutions $64 \times 64$ and $128 \times 128$, ImageNet and CelebA. We report IS/FID scores together with BPD.

	\begin{subsection}{Additional samples on CIFAR10}
				\begin{figure}[h]
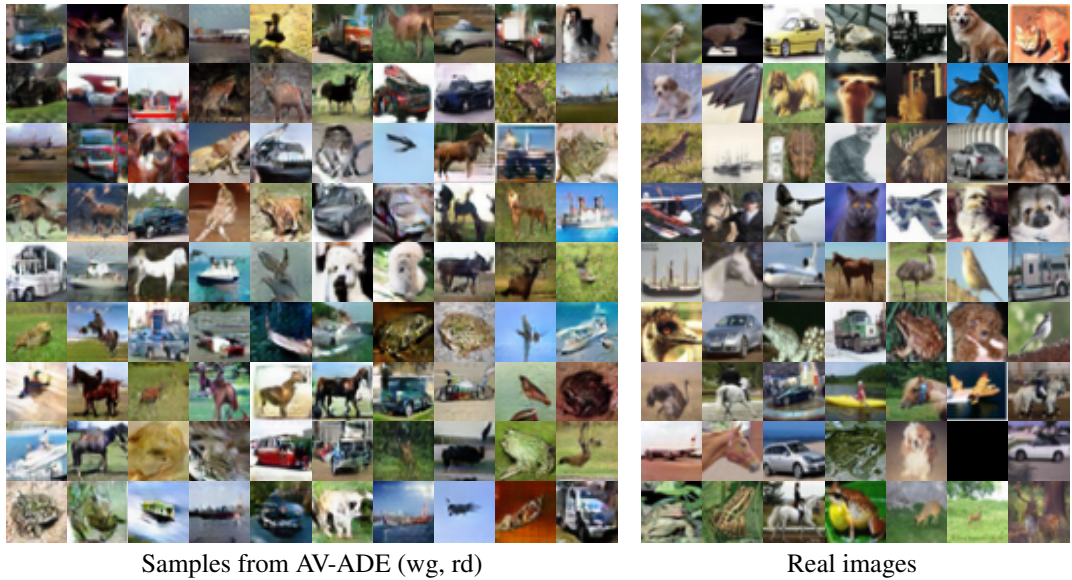

		
	\tabcolsep=2pt
	\hspace*{-0.4cm}
	\begin{tabular}{cc}
	\setlength{\tabcolsep}{0pt}
\begin{tabular}{cccccccccc} 
\includegraphics{./images/cifar_additional/cifar10/img0.png} &
\includegraphics{./images/cifar_additional/cifar10/img2.png} &
\includegraphics{./images/cifar_additional/cifar10/img3.png} &
\includegraphics{./images/cifar_additional/cifar10/img6.png} &
\includegraphics{./images/cifar_additional/cifar10/img7.png} &
\includegraphics{./images/cifar_additional/cifar10/img8.png} &
\includegraphics{./images/cifar_additional/cifar10/img9.png} &
\includegraphics{./images/cifar_additional/cifar10/img11.png} &
\includegraphics{./images/cifar_additional/cifar10/img12.png} &
\includegraphics{./images/cifar_additional/cifar10/img14.png}\\[-0.85ex]
\includegraphics{./images/cifar_additional/cifar10/img16.png} &
\includegraphics{./images/cifar_additional/cifar10/img19.png} &
\includegraphics{./images/cifar_additional/cifar10/img20.png} &
\includegraphics{./images/cifar_additional/cifar10/img21.png} &
\includegraphics{./images/cifar_additional/cifar10/img26.png} &
\includegraphics{./images/cifar_additional/cifar10/img30.png} &
\includegraphics{./images/cifar_additional/cifar10/img33.png} &
\includegraphics{./images/cifar_additional/cifar10/img34.png} &
\includegraphics{./images/cifar_additional/cifar10/img35.png} &
\includegraphics{./images/cifar_additional/cifar10/img40.png}\\[-0.85ex]
\includegraphics{./images/cifar_additional/cifar10/img41.png} &
\includegraphics{./images/cifar_additional/cifar10/img44.png} &
\includegraphics{./images/cifar_additional/cifar10/img46.png} &
\includegraphics{./images/cifar_additional/cifar10/img47.png} &
\includegraphics{./images/cifar_additional/cifar10/img48.png} &
\includegraphics{./images/cifar_additional/cifar10/img49.png} &
\includegraphics{./images/cifar_additional/cifar10/img51.png} &
\includegraphics{./images/cifar_additional/cifar10/img52.png} &
\includegraphics{./images/cifar_additional/cifar10/img53.png} &
\includegraphics{./images/cifar_additional/cifar10/img55.png}\\[-0.85ex]
\includegraphics{./images/cifar_additional/cifar10/img57.png} &
\includegraphics{./images/cifar_additional/cifar10/img59.png} &
\includegraphics{./images/cifar_additional/cifar10/img60.png} &
\includegraphics{./images/cifar_additional/cifar10/img61.png} &
\includegraphics{./images/cifar_additional/cifar10/img62.png} &
\includegraphics{./images/cifar_additional/cifar10/img63.png} &
\includegraphics{./images/cifar_additional/cifar10/img64.png} &
\includegraphics{./images/cifar_additional/cifar10/img70.png} &
\includegraphics{./images/cifar_additional/cifar10/img72.png} &
\includegraphics{./images/cifar_additional/cifar10/img73.png}\\[-0.85ex]
\includegraphics{./images/cifar_additional/cifar10/img74.png} &
\includegraphics{./images/cifar_additional/cifar10/img75.png} &
\includegraphics{./images/cifar_additional/cifar10/img76.png} &
\includegraphics{./images/cifar_additional/cifar10/img79.png} &
\includegraphics{./images/cifar_additional/cifar10/img80.png} &
\includegraphics{./images/cifar_additional/cifar10/img83.png} &
\includegraphics{./images/cifar_additional/cifar10/img84.png} &
\includegraphics{./images/cifar_additional/cifar10/img88.png} &
\includegraphics{./images/cifar_additional/cifar10/img94.png} &
\includegraphics{./images/cifar_additional/cifar10/img95.png}\\[-0.85ex]
\includegraphics{./images/cifar_additional/cifar10/img98.png} &
\includegraphics{./images/cifar_additional/cifar10/img100.png} &
\includegraphics{./images/cifar_additional/cifar10/img101.png} &
\includegraphics{./images/cifar_additional/cifar10/img102.png} &
\includegraphics{./images/cifar_additional/cifar10/img103.png} &
\includegraphics{./images/cifar_additional/cifar10/img104.png} &
\includegraphics{./images/cifar_additional/cifar10/img106.png} &
\includegraphics{./images/cifar_additional/cifar10/img108.png} &
\includegraphics{./images/cifar_additional/cifar10/img110.png} &
\includegraphics{./images/cifar_additional/cifar10/img112.png}\\[-0.85ex]
\includegraphics{./images/cifar_additional/cifar10/img114.png} &
\includegraphics{./images/cifar_additional/cifar10/img124.png} &
\includegraphics{./images/cifar_additional/cifar10/img126.png} &
\includegraphics{./images/cifar_additional/cifar10/img127.png} &
\includegraphics{./images/cifar_additional/cifar10/img129.png} &
\includegraphics{./images/cifar_additional/cifar10/img134.png} &
\includegraphics{./images/cifar_additional/cifar10/img135.png} &
\includegraphics{./images/cifar_additional/cifar10/img141.png} &
\includegraphics{./images/cifar_additional/cifar10/img144.png} &
\includegraphics{./images/cifar_additional/cifar10/img145.png}\\[-0.85ex]
\includegraphics{./images/cifar_additional/cifar10/img149.png} &
\includegraphics{./images/cifar_additional/cifar10/img151.png} &
\includegraphics{./images/cifar_additional/cifar10/img154.png} &
\includegraphics{./images/cifar_additional/cifar10/img155.png} &
\includegraphics{./images/cifar_additional/cifar10/img158.png} &
\includegraphics{./images/cifar_additional/cifar10/img163.png} &
\includegraphics{./images/cifar_additional/cifar10/img164.png} &
\includegraphics{./images/cifar_additional/cifar10/img165.png} &
\includegraphics{./images/cifar_additional/cifar10/img167.png} &
\includegraphics{./images/cifar_additional/cifar10/img169.png}\\[-0.85ex]
\includegraphics{./images/cifar_additional/cifar10/img175.png} &
\includegraphics{./images/cifar_additional/cifar10/img176.png} &
\includegraphics{./images/cifar_additional/cifar10/img177.png} &
\includegraphics{./images/cifar_additional/cifar10/img179.png} &
\includegraphics{./images/cifar_additional/cifar10/img181.png} &
\includegraphics{./images/cifar_additional/cifar10/img183.png} &
\includegraphics{./images/cifar_additional/cifar10/img185.png} &
\includegraphics{./images/cifar_additional/cifar10/img188.png} &
\includegraphics{./images/cifar_additional/cifar10/img191.png} &
\includegraphics{./images/cifar_additional/cifar10/img196.png}\\
\end{tabular} 
 &
	\setlength{\tabcolsep}{0pt}
\begin{tabular}{ccccccc} 
\includegraphics{./images/cifar_stl10/cifar10Reals_additional/img0.png} &
\includegraphics{./images/cifar_stl10/cifar10Reals_additional/img1.png} &
\includegraphics{./images/cifar_stl10/cifar10Reals_additional/img2.png} &
\includegraphics{./images/cifar_stl10/cifar10Reals_additional/img3.png} &
\includegraphics{./images/cifar_stl10/cifar10Reals_additional/img4.png} &
\includegraphics{./images/cifar_stl10/cifar10Reals_additional/img5.png} &
\includegraphics{./images/cifar_stl10/cifar10Reals_additional/img6.png}\\[-0.85ex]
\includegraphics{./images/cifar_stl10/cifar10Reals_additional/img7.png} &
\includegraphics{./images/cifar_stl10/cifar10Reals_additional/img8.png} &
\includegraphics{./images/cifar_stl10/cifar10Reals_additional/img9.png} &
\includegraphics{./images/cifar_stl10/cifar10Reals_additional/img10.png} &
\includegraphics{./images/cifar_stl10/cifar10Reals_additional/img11.png} &
\includegraphics{./images/cifar_stl10/cifar10Reals_additional/img12.png} &
\includegraphics{./images/cifar_stl10/cifar10Reals_additional/img13.png}\\[-0.85ex]
\includegraphics{./images/cifar_stl10/cifar10Reals_additional/img14.png} &
\includegraphics{./images/cifar_stl10/cifar10Reals_additional/img15.png} &
\includegraphics{./images/cifar_stl10/cifar10Reals_additional/img16.png} &
\includegraphics{./images/cifar_stl10/cifar10Reals_additional/img17.png} &
\includegraphics{./images/cifar_stl10/cifar10Reals_additional/img18.png} &
\includegraphics{./images/cifar_stl10/cifar10Reals_additional/img19.png} &
\includegraphics{./images/cifar_stl10/cifar10Reals_additional/img20.png}\\[-0.85ex]
\includegraphics{./images/cifar_stl10/cifar10Reals_additional/img21.png} &
\includegraphics{./images/cifar_stl10/cifar10Reals_additional/img22.png} &
\includegraphics{./images/cifar_stl10/cifar10Reals_additional/img23.png} &
\includegraphics{./images/cifar_stl10/cifar10Reals_additional/img24.png} &
\includegraphics{./images/cifar_stl10/cifar10Reals_additional/img25.png} &
\includegraphics{./images/cifar_stl10/cifar10Reals_additional/img26.png} &
\includegraphics{./images/cifar_stl10/cifar10Reals_additional/img27.png}\\[-0.85ex]
\includegraphics{./images/cifar_stl10/cifar10Reals_additional/img28.png} &
\includegraphics{./images/cifar_stl10/cifar10Reals_additional/img29.png} &
\includegraphics{./images/cifar_stl10/cifar10Reals_additional/img30.png} &
\includegraphics{./images/cifar_stl10/cifar10Reals_additional/img31.png} &
\includegraphics{./images/cifar_stl10/cifar10Reals_additional/img32.png} &
\includegraphics{./images/cifar_stl10/cifar10Reals_additional/img33.png} &
\includegraphics{./images/cifar_stl10/cifar10Reals_additional/img34.png}\\[-0.85ex]
\includegraphics{./images/cifar_stl10/cifar10Reals_additional/img35.png} &
\includegraphics{./images/cifar_stl10/cifar10Reals_additional/img36.png} &
\includegraphics{./images/cifar_stl10/cifar10Reals_additional/img37.png} &
\includegraphics{./images/cifar_stl10/cifar10Reals_additional/img38.png} &
\includegraphics{./images/cifar_stl10/cifar10Reals_additional/img39.png} &
\includegraphics{./images/cifar_stl10/cifar10Reals_additional/img40.png} &
\includegraphics{./images/cifar_stl10/cifar10Reals_additional/img41.png}\\[-0.85ex]
\includegraphics{./images/cifar_stl10/cifar10Reals_additional/img42.png} &
\includegraphics{./images/cifar_stl10/cifar10Reals_additional/img43.png} &
\includegraphics{./images/cifar_stl10/cifar10Reals_additional/img44.png} &
\includegraphics{./images/cifar_stl10/cifar10Reals_additional/img45.png} &
\includegraphics{./images/cifar_stl10/cifar10Reals_additional/img46.png} &
\includegraphics{./images/cifar_stl10/cifar10Reals_additional/img47.png} &
\includegraphics{./images/cifar_stl10/cifar10Reals_additional/img48.png}\\[-0.85ex]
\includegraphics{./images/cifar_stl10/cifar10Reals_additional/img49.png} &
\includegraphics{./images/cifar_stl10/cifar10Reals_additional/img50.png} &
\includegraphics{./images/cifar_stl10/cifar10Reals_additional/img51.png} &
\includegraphics{./images/cifar_stl10/cifar10Reals_additional/img52.png} &
\includegraphics{./images/cifar_stl10/cifar10Reals_additional/img53.png} &
\includegraphics{./images/cifar_stl10/cifar10Reals_additional/img54.png} &
\includegraphics{./images/cifar_stl10/cifar10Reals_additional/img55.png}\\[-0.85ex]
\includegraphics{./images/cifar_stl10/cifar10Reals_additional/img56.png} &
\includegraphics{./images/cifar_stl10/cifar10Reals_additional/img57.png} &
\includegraphics{./images/cifar_stl10/cifar10Reals_additional/img58.png} &
\includegraphics{./images/cifar_stl10/cifar10Reals_additional/img59.png} &
\includegraphics{./images/cifar_stl10/cifar10Reals_additional/img60.png} &
\includegraphics{./images/cifar_stl10/cifar10Reals_additional/img61.png} &
\includegraphics{./images/cifar_stl10/cifar10Reals_additional/img62.png}\\
\end{tabular} 
\\
	Samples from \CQFG (wg, rd) & Real images 
	\end{tabular}

		\caption{Samples from our \CQFG (wg, rd) model trained on CIFAR10 compared to real images. A significant proportion of samples can be reasonably attributed to a CIFAR10 class, even though the model was trained without labels. Our model is constrained to cover the full support of the data, which translates to diverse samples, as noted in the qualitative results illustrated here.}
		\label{fig:additional_cifar}
		\end{figure}
	\end{subsection}

	\begin{subsection}{Quantitative results on LSUN categories}
		\begin{table}[h]
		\LsunTable
		\caption{Quantitative results obtained on LSUN categories, at resolutions $64 \times 64$ and $128 \times 128$ using
		our \CQFG (wg, rd) model.}
		\label{fig:additional_lsun}
		\end{table}
	\end{subsection}

	\begin{subsection}{Samples on all Lsun datasets}
		                                                                                 				\vspace*{-1cm}
		\begin{figure}[H]
			
		\tabcolsep=2pt

	
			\label{fig:additional_sample_bed}
			\caption{Samples obtained from our \CQFG (wg, rd) model trained on LSUN bedrooms (left), compared to training images (right).}
		\end{figure}
		\vspace{-1cm}

		\begin{figure}[H]
			
		\tabcolsep=2pt
		%

			\label{fig:additional_sample_tow}
			\caption{Samples obtained from our \CQFG (wg, rd) model trained on LSUN towers (left), compared to training images (right).}
		\end{figure}

		\begin{figure}[H]
			
		\tabcolsep=2pt
		%

			\label{fig:additional_sample_bridge}
			\caption{Samples obtained from our \CQFG (wg, rd) model trained on LSUN bridges (left), compared to training images (right).}
		\end{figure}

		\begin{figure}[H]
			
		\tabcolsep=2pt
		%

			\label{fig:additional_sample_kitchen}
			\caption{Samples obtained from our \CQFG (wg, rd) model trained on LSUN kitchen (left), compared to training images (right).}
		\end{figure}

		\begin{figure}[H]
			
		\tabcolsep=2pt
		%

			\label{fig:additional_sample_church}
			\caption{Samples obtained from our \CQFG (wg, rd) model trained on LSUN churches (left), compared to training images (right).}
		\end{figure}

		\begin{figure}[H]
			
		\tabcolsep=2pt
		%

			\label{fig:additional_sample_living}
			\caption{Samples obtained from our \CQFG (wg, rd) model trained on LSUN living rooms (left), compared to training images (right).}
		\end{figure}

		\begin{figure}[H]
			
		\tabcolsep=2pt
		%

			\label{fig:additional_sample_dining}
			\caption{Samples obtained from our \CQFG (wg, rd) model trained on LSUN dining rooms (left), compared to training images (right).}
		\end{figure}

		\begin{figure}[H]
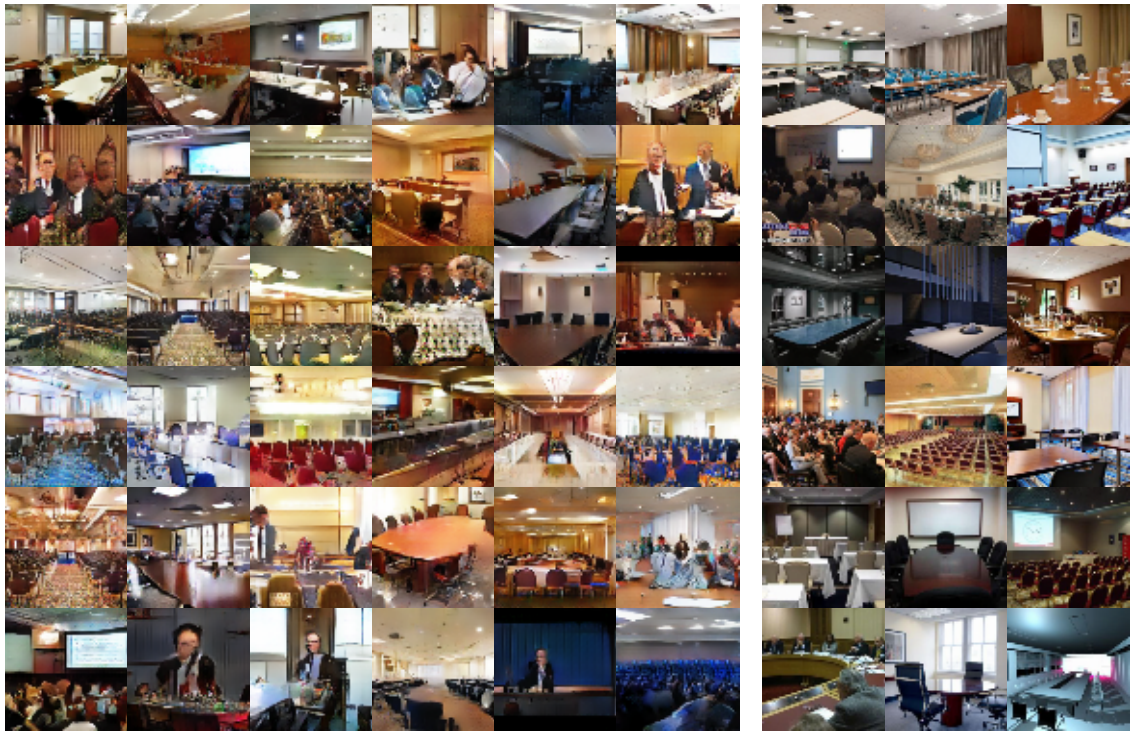

			
		\tabcolsep=2pt
		%

			\label{fig:additional_sample_conference}
			\caption{Samples obtained from our \CQFG (wg, rd) model trained on LSUN conference rooms (left), compared to training images (right).}
		\end{figure}

		\begin{figure}[H]
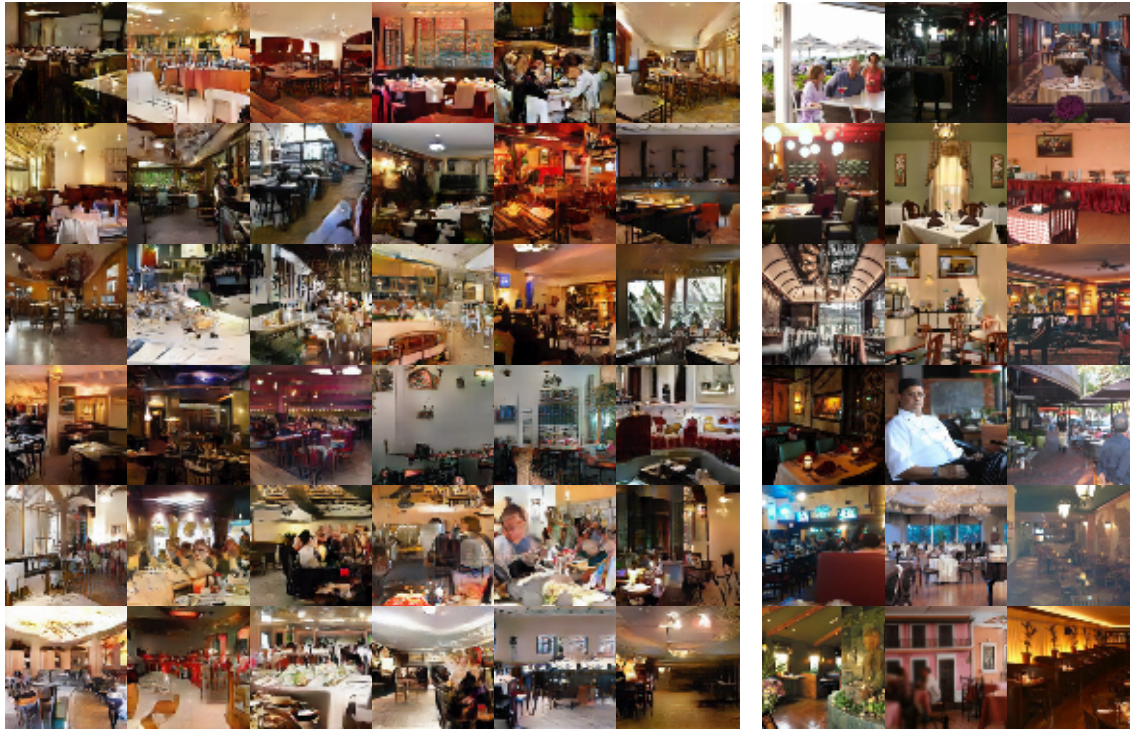

			
		\tabcolsep=2pt
		%

			\label{fig:additional_sample_restaurant}
			\caption{Samples obtained from our \CQFG (wg, rd) model trained on LSUN restaurants (left), compared to training images (right).}
		\end{figure}

	\end{subsection}

	\begin{subsection}{Additional samples on STL10}
		\begin{figure}[H]
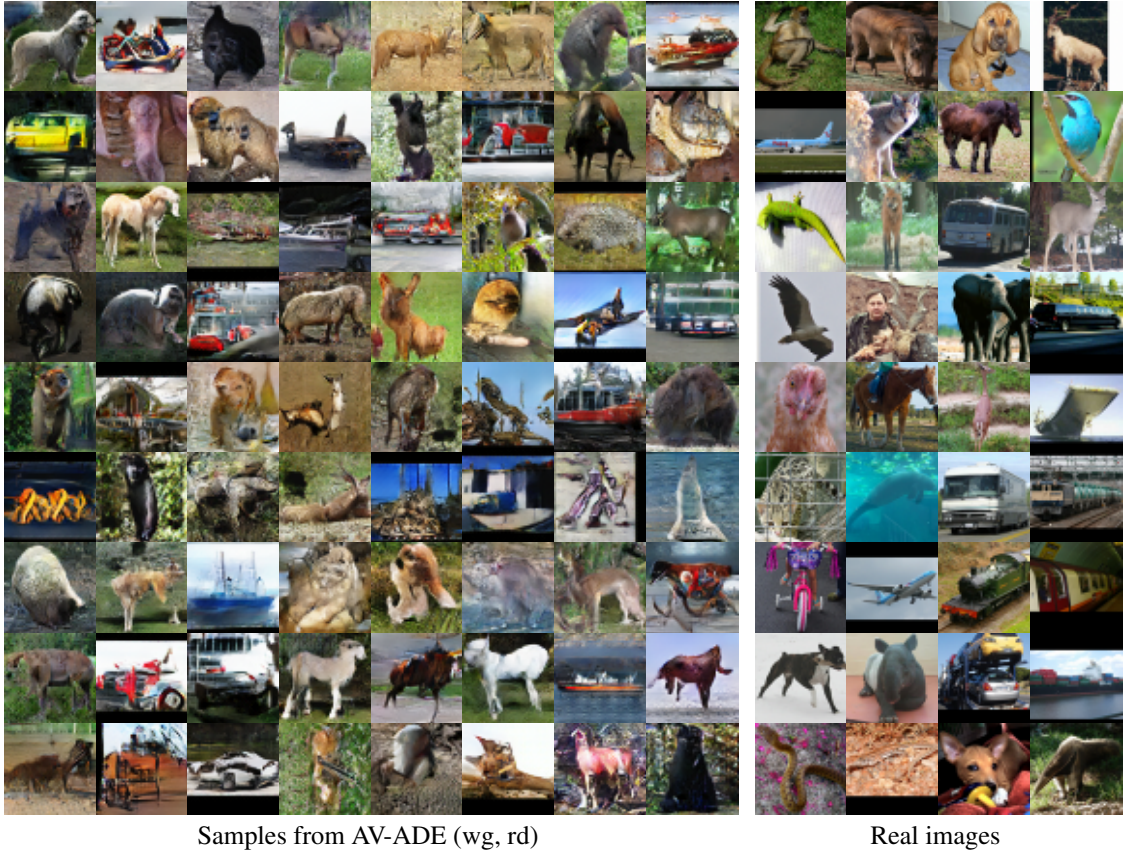

		
	\tabcolsep=2pt
	\hspace*{-0.4cm}
	%
 
\\
		Samples from \CQFG (wg, rd) & Real images 
	\end{tabular}

		\caption{Samples from our \CQFG (wg, rd) model trained on STL10 compared to real images. Training was done without any label conditioning.}
		\label{fig:additional_stl}
		\end{figure}
	\end{subsection}

\label{app:samples}
			\end{section}

\begin{section}{Model refinements}
\label{app:implementation}

	\begin{subsection}{Top-down sampling of hierarchical latent variables}
		\label{app:top_down}
		Flexible priors and posteriors for the variational autoencoder model
		can be obtained by sampling hierarchical latent variables at different layers 
		in the network. In the generative model $\q$, latent variables $\bold{z}$ can be split into
		$L$ groups, each one at a different layer, and the density over $\bold{z}$ can be written as:
		\begin{equation}
		q(\bold{z}) = q(\bold{z}_L) \prod_{i=1}^{L-1}q(\bold{z}_i|\bold{z}_{i+1}).
		\end{equation}
		Additionally, to allow the chain of latent variables to be sampled in the same order when encoding-decoding 
		and when sampling, top-down sampling is used, as proposed in \citet{ladderVAE,hierarchiVAE,iaf}.
		With top-down sampling, the encoder (symmetric to the decoder) extracts deterministic features $h_i$ at different levels as the image is being encoded, constituting
		the bottom-up deterministic pass.
		While decoding the image, these previously extracted deterministic features $h_i$ are used for top-down
		sampling and help determining the posterior over latent variables at different depths in the decoder.
		These posteriors are also conditioned on the latent variables sampled at lower feature resolutions, using normal densities as follows:
		\begin{eqnarray}
			q_{\phi}(\bold{z}_1 | x) &=& \mathcal{N}(\bold{z}_1 | \mu_1(x, h_1), \sigma^2_1(x, h_1)), \\
			q_{\phi}(\bold{z}_i | \bold{z}_{i-1}) &=& \mathcal{N}(\bold{z}_i | \mu_i(x, \bold{z}_{i-1}, h_{i-1}), \sigma^2_i(x, \bold{z}_{i-1}, h_{i-1})).
		\end{eqnarray}
		This constitutes the stochastic top-down pass. We refer the reader to \citet{ladderVAE,hierarchiVAE,iaf} for more detail.
	\end{subsection}

	\begin{subsection}{Inverse autoregressive flow}
		\label{app:iaf_short}
		To increase the flexibility of posteriors used over latent variables in variational inference,
		\citet{iaf} proposed a type of normalizing flow called inverse autoregressive flow (IAF).
		The main benefits of this normalizing flow are its scalability to high dimensions, and its ability to leverage 
		autoregressive neural network, such as those introduced by \citet{pixrnn}. 
		First, a latent variable vector is sampled using the reparametrization trick \cite{vae}: 
		\begin{equation}
		\epsilon \sim \mathcal{N}(0, I), \\
		z_0 = \mu_0 + \sigma_0 \epsilon. 
		\end{equation}
		Then, mean and variance parameters $\mu_1$ and $\sigma_1$ are computed as functions of $z_0$ using 
		autoregressive models, and a new latent variable $z_1$ is obtained:
		\begin{equation}
		z_1 = \mu_1(z_0) + \sigma_1(z_0) z_0.
		\end{equation}
		Since $\sigma_1$ and $\mu_1$ are implemented by autoregressive networks, the Jacobian $\frac{dz_1}{dz_0}$ is
		triangular with the values of $\sigma_1$ on the diagonal, and the density under the new latent variable remains efficient to compute.
		This transformation can be repeated an arbitrary number of times for increased flexibility in theory, but in practice a single step
		is used.
	\end{subsection}

	\begin{subsection}{Gradient penalty}
		\label{sec:GradPen}
		A body of work on generative adversarial networks centers around the idea of regularizing the discriminator by enforcing 
		Lipschitz continuity, for instance by \citet{sngan, wgan, gulrajani17nips, centeredwgp}.
		In this work, we use the approach of \citet{gulrajani17nips}, that enforces the Lipschitz constraint with a gradient penalty term added to the loss:
		\begin{equation}
		\mathcal{L}_{Grad} = \lambda + \mathbb{E}_{\hat{x}}[(|| \Delta_{\hat{x}} D(\hat{x})||_2 - 1)^2],
		\end{equation}
		where $\hat{x}$ is obtained by interpolating between real and generated data:
		\begin{eqnarray}
			\epsilon &\sim& U_{[0,1]},\\
			\hat{x} &=& \epsilon x + (1 - \epsilon) \tilde{x}.
		\end{eqnarray}
		We add this term to the loss used to train the discriminator that yields our quality driven criterion. 
	\end{subsection}
\end{section}

\begin{section}{Implementation details}
		\mypar{Architecture and training hyper-parameters}
	We used Adamax~\citep{adam} with learning rate 0.002, $\beta_1=0.9$, $\beta_2=0.999$ for all experiments.
	The convolutional architecture of the models used is described in \tab{nvp_d_arch} and \tab{nvp_vae_arch}. All CIFAR-10
	experiments use batch size 64, other experiments in high resolution use batch
	size 32. To stabilize the adversarial training we use the
	gradient penalty~\citep{gulrajani17nips} with coefficient 100, and 1 discriminator
	update per generator update. We experimented with different weighting coefficients between
	the two loss components, and found that values in the range $10$ to $100$ on the adversarial component
	work best in practice. No significant influence on the final performance of the model
	is observed in this range, though the training dynamics in early training are improved with higher values. With values significantly smaller 
	than $10$, discriminator collapse was observed in a few isolated cases. All experiments reported here use coefficient $100$.

	For experiments with hierarchical latent variables, we use $32$ of them per layer. 
	In the generator we use ELU nonlinearity, in discriminator with residual blocks we use
	ReLU, while in simple convolutional discriminator we use leaky ReLU with slope 0.2.

	Unless stated otherwise we use three Real-NVP layers with a single scale and two residual blocks 
	that we train only with the likelihood loss. 
	 Regardless of the number of scales, the VAE decoder always outputs a tensor of the same dimension as the target image, which is then fed to the Real-NVP layers. As in the reference implementations, we use both batch normalization and weight normalization in Real-NVP and only weight normalization in IAF. We use the reference implementations of IAF and Real-NVP released by the authors.

		\begin{table}[htb]
	  \centering
	  \subfloat{{
	  \begin{tabular}{c}
	    Discriminator \\
	    \hline
	    conv $3\times 3$, 16\\
	    \hline
	    ResBlock 32 \\
	    \hline
	    ResBlock down 64 \\
	    \hline
	    ResBlock down 128 \\
	    \hline
	    ResBlock down 256 \\
	    \hline
	    Average pooling\\
	    \hline
	    dense 1
	  \end{tabular}}\label{tab:nvp_d_arch}}
	  \subfloat{{
	  \begin{tabular}{c}
	    Generator \\
	    \hline
	    conv $3\times 3$, 16\\
	    \hline
	    IAF block 32 \\
	    \hline
	    IAF block down 64 \\
	    \hline
	    IAF block down 128 \\
	    \hline
	    IAF block down 256 \\
	    \hline
	    $h \sim \mathcal{N}(0; 1)$\\
	    \hline
	    IAF block up 256 \\
	    \hline
	    IAF block up 128 \\
	    \hline
	    IAF block up 64 \\
	    \hline
	    IAF block 32 \\
	    \hline
	    conv $3\times 3$, 3
	  \end{tabular}}\label{tab:nvp_vae_arch}}
	\caption{Residual architectures for experiments from \sect{refinements} and \tab{nvp}}
	\end{table}
\end{section}

\begin{section}{On the Inception Score and the Fr\'echet inception distance}

\label{app:AppEval}
Quantitative evaluation of Generative Adversarial Networks is challenging, in part due to the absence of log-likelihood.
Inception score (IS) and Fr\'echet Inception distance (FID) are two measures proposed 
by \cite{salimans16nips} and \cite{fid} respectively, to automate the qualitative evaluation of samples. Though imperfect, 
they have been shown to correlate well with human judgement in practice,
and it is standard in GAN literature to use them for quantitative evaluations. 
These metrics are also sensitive to coverage. Indeed, any metric evaluating quality only would be degenerate,
as collapsing to the mode of the distribution would maximize it. However, in practice both metrics are much more
sensitive to quality than to support coverage, as we evaluate below.

	\textbf{Inception score (IS)}~\citep{salimans16nips} is a statistic of the generated images, based on an external deep network trained for classification on ImageNet. It is given by:
\begin{equation}
	IS(\q) = \exp\left(\mathbb{E}_{x \sim \q} \mathcal{D}_\textrm{KL}\left(p\left(y|x\right) || p(y)\right)\right),
\end{equation}
where $x \sim \q$ is sampled from the generative model, $p(y|x)$ is the conditional class distribution obtained by 
applying the pretrained classification network to the generated images, and $p(y) = \int_x p(y|x) \q(x)$ is the class marginal over the generated images.

\textbf{Fr\'{e}chet Inception distance (FID)}~\cite{fid} compares the distributions of Inception embeddings
\ie, activations from the penultimate layer of the Inception network, of real
($p_r(\mathbf{x})$) and generated ($p_g(\mathbf{x})$) images. Both these
distributions are modeled as multi-dimensional Gaussians parameterized by their
respective mean and covariance. The distance measure is defined between the two
Gaussian distributions as:
\begin{equation}\label{eqn:fid}
  d^2((\mathbf{m}_r, \mathbf{C}_r), (\mathbf{m}_g, \mathbf{C}_g)) =
  \norm{\mathbf{m}_r - \mathbf{m}_g}^2 +\\ \tr (\mathbf{C}_r + \mathbf{C}_g - 2(\mathbf{C}_r \mathbf{C}_g)^\frac{1}{2}),
\end{equation}
where $(\mathbf{m}_r, \mathbf{C}_r)$, $(\mathbf{m}_g, \mathbf{C}_g)$ denote the
mean and covariance of the real and generated image distributions respectively.

\textbf{Practical use.}
IS and FID correlate predominantly with the quality of samples.
In GAN literature, for instance \citet{sngan}, they are considered to correlate well with human judgement of quality.
An empirical indicator of that is that state-of-the art likelihood-based models have very low IS/FID scores despite having good coverage, which shows that the low quality of their samples impacts IS/FID 
more heavily than their coverage performance.
Conversely, state-of-the art adversarial models have high IS/FID scores, despite suffering from mode dropping (which strongly degrades BPD). So the score is determined mainly by the high quality of their samples.
This is especially true when identical architectures and training budget are considered, as in our first experiment in \sect{ablation}.

\begin{table}[htb]
		\begin{center}
			\begin{tabular}{lcc}
				\toprule
 				Split size & IS & FID \\
				\midrule
				50k (full) & $11.3411$ & $0.00$ \\
				40k & $11.3388$ & $0.13$ \\
				30k & $11.3515$ & $0.35$ \\
				20k & $11.3458$ & $0.79$ \\
				10k & $11.3219$ & $2.10$ \\
				5k & $11.2108$ & $4.82$ \\
				2.5k & $11.0446$ & $10.48$ \\
				\bottomrule
			\end{tabular}
	    \caption{IS and FID scores obtained by the ground truth when progressively dropping parts of the dataset. The metrics are largely 
			insensitive to removing most of the dataset, unlike BPD. For reference, a reasonable GAN could get around $8$ IS and $20$ FID.}
	    \label{tab:eval_scores}
	\end{center}
\end{table}
\end{section}

To obtain a quantitative estimation of how much entropy/coverage impacts the IS and FID measures, 
we evaluate the scores obtained by random subsamples of the dataset, 
such that the quality is unchanged but coverage progressively degrades (see details of the scores below).
\tab{eval_scores} shows that when using the full set of images (50k) the FID is $0$ as the distributions are identical.
Notice that as the number of images decreases, IS is very stable (it can even increase, but by very low increments that fall below statistical noise, with a typical standard deviation of $0.1$).
This is because the entropy of the distribution is not strongly impacted by subsampling, even though coverage is.
FID is more sensitive, as it behaves more like a measure of coverage (it compares the two distributions).
Nonetheless, the variations remain extremely low even when dropping most of the dataset. For instance, when removing $80\%$ of the dataset (\ie., using 10k images),
FID is at $2.10$, to be compared with typical GAN/\CQFG values that are around 20. These measurements demonstrate that IS and FID scores are heavily dominated by the quality of images.
From this, we conclude that IS and FID can be used as reasonable proxies to asses sample quality, even though they are also slightly influenced by coverage.
One should bear in mind, however, that a small increase in these scores may come from better coverage rather than improved sample quality.

\begin{section}{Other evaluation metrics}
\label{app:prdKtrain}

\subsection{Evaluation using samples as training data for a discriminator}

We also evaluate our approach using the two metrics recently proposed by \citet{Ktest}.
This requires a class-conditional version of \CQFG. 
To address the poor compatibility of conditional batch-normalization with VAEs,
we propose conditional weight normalization (CWN), see below for details.
Apart from this adaptation, the architecture is the same as in \sect{ablation}. 

The first metric, GAN-test, is obtained by training a classifier on natural image data
and evaluating it on generated samples. It is sensitive to sample quality only.
Our \CQFG model obtains a slightly higher GAN-test score, suggesting comparable sample quality, which is in line with the results in \sect{ablation}.
\begin{table}[H]
	\footnotesize
	\setlength{\tabcolsep}{3pt}
		\begin{center}
			\begin{tabular}{lcc}
				\toprule
				model & GAN-test (\%) & GAN-train (\%)\\
				\midrule
				GAN   & 71.8     & 29.7 \\
				\CQFG  & \textbf{76.9}     & \textbf{73.4}\\
				\midrule
				DCGAN$^\dag$ & 58.2 & 65.0 \\
				\bottomrule
			\end{tabular}
		\end{center}
		\caption{GAN-test and GAN-train measures for class conditional CQFG and GAN models on CIFAR-10. 
		The performance of the DCGAN$^\dag$  model, though not directly comparable, is provided as a reference point.}
			\end{table}

The second metric, GAN-train, is obtained by training a classifier on generated samples and evaluating it on natural images.
Having established similar GAN-test performance, this demonstrates improved sample diversity of the \CQFG model and shows that the coverage-driven
training improves the support of the learned model. 

\mypar{Class conditioning with conditional weight-normalization} To perform this evaluation we develop a class conditional version of our \CQFG model.
The discriminator is conditioned using the class conditioning introduced by \citet{cgan}. 
GAN generators are typically made class-conditional using conditional batch normalization \cite{de2017modulating,dumoulin2017learned},
however batch normalization is known to be detrimental in VAEs~\cite{iaf}, as we verified in practice.
To address this issue, we propose conditional weight normalization (CWN).
As in weight normalization \cite{weightnorm}, we separate the training of
the scale and the direction of the weight matrix. Additionally, the scaling factor $g(y)$ of the weight matrix $\bf v$ is conditioned on the class label $y$:
\begin{equation}
    \mathbf{w} = \frac{g(y)}{\norm{\mathbf{v}}}\mathbf{v},
\end{equation}
We also make the network biases conditional on the class label.
Otherwise, the architecture is the same one used for the experiments in  \sect{ablation}.

\subsection{Model evaluation using precision and recall}
	In this section, we evaluate our models using the precision and recall procedure of~\citet{prdEval}. 
	This evaluation is relevant as it seeks to evaluate coverage of the support and the quality of the samples separately,
	rather than aggregating them into a single score.
Intuitively, the precision measures the sample quality of the model, while the recall measures to what extent the model covers the support of the target distribution.
The precision-recall metrics are determined based on a clustering of \emph{Pool3} features extracted from training images and generated images using a pre-trained Inception network. 
The resulting histograms over cluster assignments are then used to assess the similarity of the distributions.

	\begin{figure}[H]
				\includegraphics[width=0.7\textwidth]{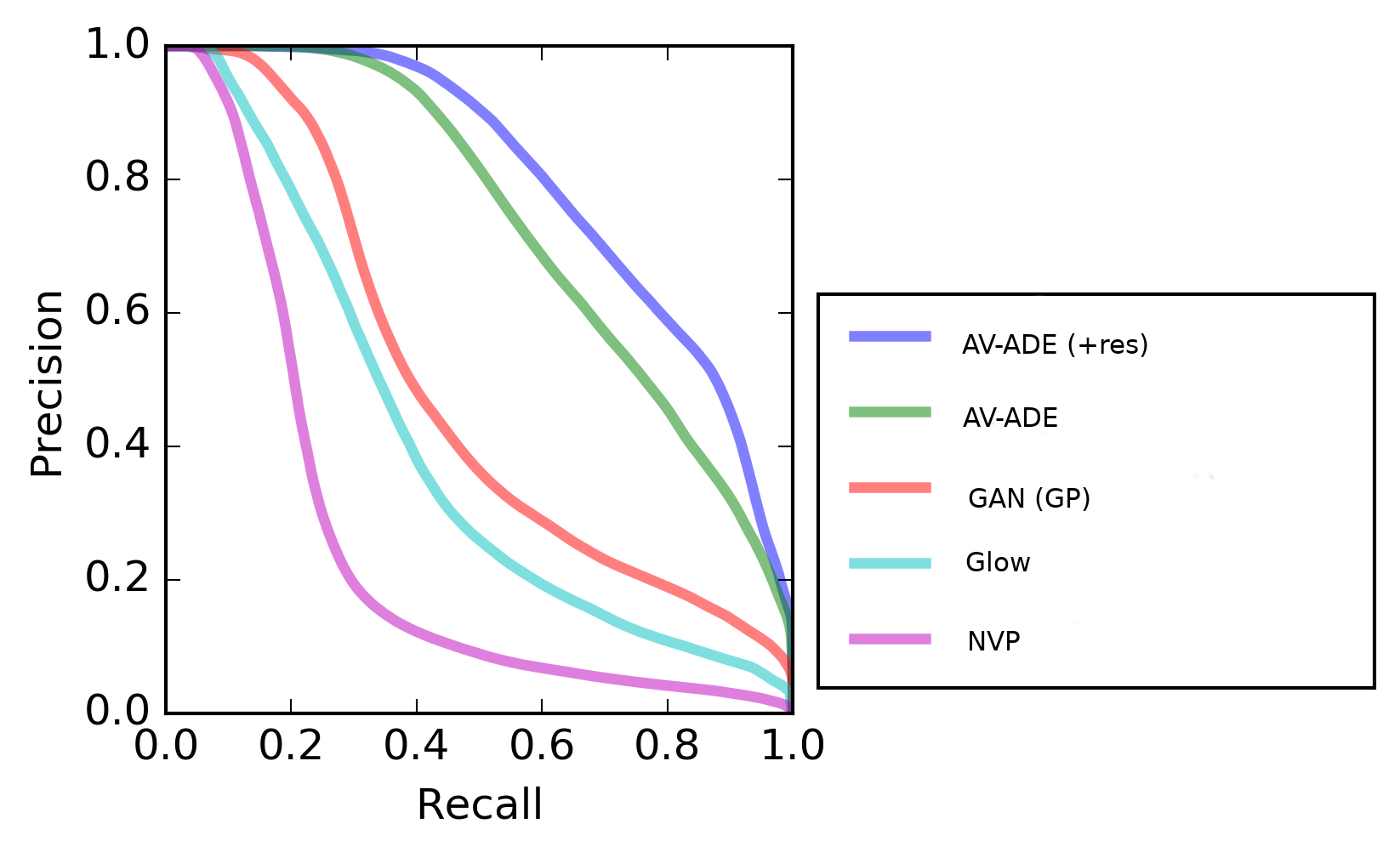} 
		\caption{Precision-recall curves using the evaluation procedure of~\citep{prdEval}.}
		\label{fig:prdCurve}
	\end{figure}
\fig{prdCurve} presents the evaluation of our models in \sect{ablation}, as
well as the Glow and Real-NVP models, using the official code provided online by the
authors at \url{https://github.com/msmsajjadi/precision-recall-distributions}.
Our \CQFG model obtains a better area under curve (AUC) than the GAN baseline,
and model refinements improve AUC further. For comparison, Glow and Real-NVP have
lower AUC than both GAN and our models.
\end{section}

\begin{section}{Qualitative influence of the feature space flexibility}\label{app:nvpPurelikeSamples}
\label{app:features}

We experiment with different architectures to implement the invertible mapping used to build the feature space as presented in \sect{ADE}. 
To assess the impact of the expressiveness of the invertible model on
the behavior of our framework, we modify various standard parameters of the architecture.
Popular invertible models such as Real-NVP~\citep{realnvp}
readily offer the possibility of extracting latent representation at several scales,
separating global factors of variations from low level detail. Thus, we experiment with varying number of scales.
Another way of increasing the flexibility of the model is to change the number of residual blocks used
in each invertible layer. Note that all the models evaluated so far in the main body of the paper are based on
a single scale and two residual blocks, except the one denoted with (s2).
In addition to our \CQFG models, we also compare with similar models trained with maximum likelihood estimation (V-ADE).
Models are first trained with maximum-likelihood estimation, then with both coverage and quality driven criteria.

The results in \tab{nvp} show that factoring out features at two scales rather than one is helpful in terms of BPD. For the \CQFG models, however, IS
and FID deteriorate with more scales, and so a tradeoff between must be struck.
For the V-ADE models, the visual quality of samples also improves when using multiple scales, as reflected in better IS and FID scores. 
Their quality, however, remains far worse than those produced with the coverage and quality training used for the \CQFG models.
Samples in the maximum-likelihood setting are provided in \fig{nvpAblation}.
With three or more scales, models exhibit symptoms of overfitting: train BPD keeps decreasing while test BPD starts increasing, and IS and FID also degrade.

     \begin{table}[htb]
		\begin{center}
		 \subfloat[\CQFG models]{
        {
			\begin{tabular}{ccccc}
				\toprule
				Scales     & Blocks    &  BPD $\downarrow$ & IS $\uparrow$ & FID $\downarrow$\\
				\midrule
				 1          & 2               &3.77   & \textbf{7.9}   & \textbf{20.1 }\\
				 2          & 2               & 3.48          & 6.9       & 27.7   \\
				 2          & 4               & \textbf{3.46} & 6.9       & 28.9   \\
				 3          & 3               & 3.49          & 6.5       & 31.7   \\
				 \bottomrule
			\end{tabular}
		        }\label{tab:nvp_full}
      }      \hfill
      \subfloat[V-ADE models]{
	{
			\begin{tabular}{ccccc}
				\toprule
				Scales     & Blocks    & BPD $\downarrow$ & IS $\uparrow$ & FID $\downarrow$\\
				\midrule
				 1          & 2               & 3.52       & 3.0          & 112.0   \\
				 2          & 2               & \textbf{3.41}       & \textbf{4.5}   & 85.5    \\
				 3          & 2               & 3.45       & 4.4       & \textbf{78.7}   \\
				 4          & 1               & 3.49       & 4.1       & 82.4   \\
				 				 \bottomrule
        			\end{tabular}
        }\label{tab:nvp_purelike}
      }     
      \end{center}
      \caption{Evaluation on CIFAR-10 of different architectures of the invertible layers of the model. 
			}
			\label{tab:nvp} 
    \end{table}

    \begin{figure}[bth]
	    \begin{tabular}{cc}
		    \hspace*{-0.3cm} \setlength{\tabcolsep}{0pt}
\begin{tabular}{ccccc} 
\includegraphics{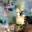} &
\includegraphics{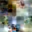} &
\includegraphics{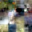} &
\includegraphics{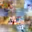} &
\includegraphics{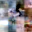} \\[-0.85ex]
\includegraphics{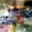} &
\includegraphics{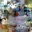} &
\includegraphics{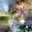} &
\includegraphics{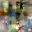} &
\includegraphics{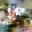} 
\end{tabular} 
 & \hspace*{-0.3cm} \setlength{\tabcolsep}{0pt}
\begin{tabular}{ccccc} 
\includegraphics{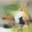} &
\includegraphics{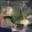} &
\includegraphics{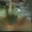} &
\includegraphics{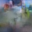} &
\includegraphics{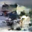} \\[-0.85ex]
\includegraphics{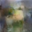} &
\includegraphics{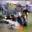} &
\includegraphics{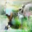} &
\includegraphics{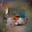} &
\includegraphics{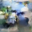} 
\end{tabular} 
 \\
        		No $f_\psi$ & $f_\psi$ 1 scale\\
		    \hspace*{-0.3cm} \setlength{\tabcolsep}{0pt}
\begin{tabular}{ccccc} 
\includegraphics{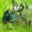} &
\includegraphics{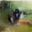} &
\includegraphics{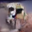} &
\includegraphics{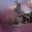} &
\includegraphics{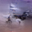} \\[-0.85ex]
\includegraphics{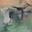} &
\includegraphics{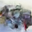} &
\includegraphics{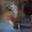} &
\includegraphics{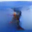} &
\includegraphics{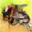} 
\end{tabular} 
   & \hspace*{-0.3cm} \setlength{\tabcolsep}{0pt}
\begin{tabular}{ccccc} 
\includegraphics{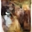} &
\includegraphics{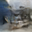} &
\includegraphics{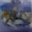} &
\includegraphics{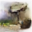} &
\includegraphics{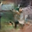} \\[-0.85ex]
\includegraphics{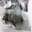} &
\includegraphics{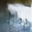} &
\includegraphics{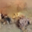} &
\includegraphics{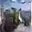} &
\includegraphics{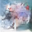}
\end{tabular} 
 \\
        		$f_\psi$ 2 scales & $f_\psi$ 3 scales \\
	    \end{tabular}
	\caption{Samples obtained using VAE models trained with MLE (\tab{nvp_purelike}) showing
        qualitative influence of multi-scale feature space. The models include one
        without invertible decoder layers, and with real-NVP layers using one, two and three
        scales. The samples illustrate the impact of using invertible real-NVP layers in these autoencoders.}
    	\label{fig:nvpAblation}
    \end{figure}

\end{section}

\begin{section}{Visualisations of reconstructions}\label{app:Reconstruct}

We display reconstructions obtained by encoding and then decoding ground truth 
images with our models (\CQFG from \sect{ablation}) in \fig{Reconstructions}. 
As is typical for expressive variational autoencoders, real images and their reconstructions cannot be distinguished visually.

    \begin{figure}[htb!]
            \begin{tabular}{cc}
		 				 \hspace*{-0.3cm} \includegraphics[scale=0.74, trim={0 3.4cm 6.8cm 0}, clip]{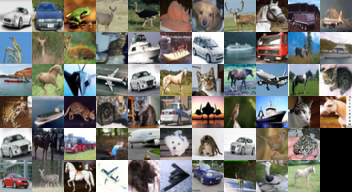} &  \hspace*{-0.3cm} \includegraphics[scale=0.75, trim={0 3.4cm 6.8cm 0}, clip]{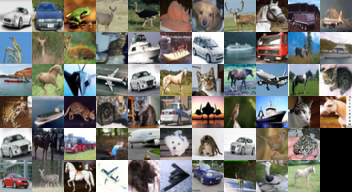}\\
		Real image  & \CQFG reconstruction \\
        	                    \end{tabular}
          \caption{Real images and their reconstructions with the \CQFG models.}
    	\label{fig:Reconstructions}
    \end{figure}
\end{section}

\begin{section}{Coverage and Quality driven training algorithm}
\label{app:algorithm}

\begin{algorithm}[H]
\caption{\small Coverage and Quality driven training for our \CQFG model.}
\begin{algorithmic}
\label{alg:CQFG}
\FOR{number of training steps}
      \STATE{$\bullet$ Sample $m$ real images $\{\bm{x}^{(1)}, \dots, \bm{x}^{(m)} \}$ from $\p$, approximated by the dataset.}

    \STATE{$\bullet$ Map the real images to feature space $\{f(\bm{x}^{(1)}), \dots, f(\bm{x}^{(m)}) \}$ using the invertible transformation $f$.}

    \STATE{$\bullet$ Encode the feature space vectors using the VAE encoder and get parameters for the posterior $q_{\phi}(\bm{z}|f(\bm{x}))$.}

    \STATE{$\bullet$ Sample $m$ latent variable vectors, $\{ \hat{\bm{z}}^{(1)}, \dots, \hat{\bm{z}}^{(m)} \}$ from the posterior $q_{\phi}(z|x)$, and $m$ latent variable vectors $\{ \tilde{\bm{z}}^{(1)}, \dots, \tilde{\bm{z}}^{(m)} \}$ from the VAE prior $p_{\theta}(z)$}

    \STATE{$\bullet$ Decode both sets of latent variable vectors using the VAE decoder into the means of conditional Gaussian distributions, $\{ \mu(\hat{\bm{z}})^{(1)}, \dots, \mu(\hat{\bm{z}})^{(m)} \}$ and $\{ \mu(\tilde{\bm{z}})^{(1)}, \dots, \mu(\tilde{\bm{z}})^{(m)} \}$}

    \STATE{$\bullet$ Sample from the Gaussian densities obtained, $\{ \mathcal{N}(.| \mu(\hat{\bm{z}})^{(i)}, \sigma I_n) \}_{i \leq m}$ and $\{ \mathcal{N}(.| \mu(\tilde{\bm{z}})^{(i)}, \sigma I_n) \}_{i \leq m}$, which yields reconstructions in feature space $\{ \widehat{f(\bm{x})}^{(i)} \}_{i \leq m}$ and samples in feature space $\{ \widetilde{f(\bm{x})}^{(i)} \}_{i \leq m}$}

	\STATE{$\bullet$ Map the samples and reconstructions back to image space using the inverse of the invertible transformation $ f^{-1}$ which yields reconstructions $\{ \hat{\bm{x}}^{(i)} \}_{i \leq m}$ and samples $\{ \tilde{\bm{x}}^{(i)} \}_{i \leq m}$}

	\STATE{$\bullet$ Compute $\mathcal{L}_\textrm{C}(\q)$ using ground truth images $\{\bm{x}^{(i)} \}_{i \leq m}$ and their reconstructions $\{\bm{\hat{x}}^{(i)} \}_{i \leq m}$}

	\STATE{$\bullet$ Compute $\mathcal{L}_\textrm{Q}(\q)$ by feeding the ground truth images $\{\bm{x}^{(i)} \}_{i \leq m}$ together with the sampled images $\{\bm{\tilde{x}}^{(i)} \}_{i \leq m}$ to the discriminator}

	\STATE{$\bullet$ Optimize the discriminator by gradient descent to bring $\mathcal{L}_Q$ closer to $\mathcal{L}_\textrm{Q}^*$}

	\STATE{$\bullet$ Optimize the generator by gradient descent to minimize $\mathcal{L}_\textrm{Q} + \mathcal{L}_\textrm{C}$}
                                                                                                                                                \ENDFOR
  \end{algorithmic}
\end{algorithm}

\end{section}

\end{document}